\documentclass[letterpaper, 10 pt, conference]{ieeeconf} 

\IEEEoverridecommandlockouts  
\usepackage{graphicx}
\usepackage{amsmath,amssymb} 
\usepackage{color}
\usepackage{multicol}
\usepackage{bm}
\usepackage{url}
\usepackage{multirow}
\usepackage{times}
\usepackage{gensymb}
\usepackage{cite}
\usepackage[bookmarks=false]{hyperref}
\hypersetup{
    colorlinks=true,     
    urlcolor=cyan,
}

\usepackage[font=footnotesize,labelformat=simple]{subcaption}
\DeclareCaptionLabelSeparator{periodspace}{.\quad}
\definecolor{rColor}{rgb}{0, 0, 1}

\definecolor{bColor}{rgb}{1, 0, 0}

\graphicspath{ {./figures/}} 

\title{\LARGE \bf Dense 3D Reconstruction for Visual Tunnel Inspection using Unmanned Aerial Vehicle}
\author{Ramanpreet~Singh~Pahwa$^{1}$,
		Kennard~Yanting~Chan$^{2}$,
		Jiamin~Bai$^{1}$,
		Vincensius~Billy~Saputra$^{1}$,  \\
		Minh~N.~Do$^{3}$, and
		Shaohui~Foong$^{4}$%
\thanks{$^{1}$R.~S.~Pahwa, J.~Bai, and  V.~B.~Saputra are with Institute for Infocomm Research (I$^2$R), A*STAR, Singapore. }%
\thanks{$^{2}$K.~Y.~Chan is with Nanyang Technological University (NTU), Singapore. }
\thanks{$^{2}$Minh~N.~Do is with University of Illinois at Urbana-Champaign (UIUC), USA. }
\thanks{$^{4}$S.~Foong is with Singapore University of Technology and Design (SUTD), Singapore.}%
}

\begin{document}
\maketitle
\thispagestyle{empty}
\pagestyle{empty}
\begin{abstract}
Advances in Unmanned Aerial Vehicle (UAV) opens venues for application such as tunnel inspection. Owing to its versatility to fly inside the tunnels, it can quickly identify defects and potential problems related to safety. However, long tunnels, especially with repetitive or uniform structures pose a significant problem for UAV navigation. Furthermore, post-processing visual data from the camera mounted on the UAV is required to generate useful information for the inspection task. In this work, we design a UAV with a single rotating camera to accomplish the task. Compared to other platforms, our solution can fit the stringent requirement for tunnel inspection, in terms of battery life, size and weight. While the current state-of-the-art can estimate camera pose and 3D geometry from a sequence of images, they assume large overlap, small rotational motion, and many distinct matching points between images. These assumptions severely limit their effectiveness in tunnel-like scenarios where the camera has erratic or large rotational motion, such as the one mounted on the UAV. This paper presents a novel solution which exploits Structure-from-Motion, Bundle Adjustment, and available geometry priors to robustly estimate camera pose and automatically reconstruct a fully-dense 3D scene using the least possible number of images in various challenging tunnel-like environments. We validate our system with both Virtual Reality application and experimentation with a real dataset. The results demonstrate that the proposed reconstruction along with texture mapping allows for remote navigation and inspection of tunnel-like environments, even those which are inaccessible for humans.
\end{abstract}

\section{Introduction}
Recent advances in low-powered Unmanned Aerial Vehicles (UAVs) and drones have enabled them as acquisition tools for $3$D reconstruction as well as scene visualization, anomaly and fault detection, and Dense Surface Models (DSM) creation. Although image stitching, Structure-from-Motion (SfM), Simultaneous Localization and Mapping (SLAM), and $3$D reconstruction are mature topics in the Computer Vision community \cite{StereoScan, 3DModelling,dense_reconstruction_fly, raman_tcsvt_3D_prop, raman_apsipa_3D_prop}, most of these techniques use a depth camera or focus on sparse and semi-dense reconstruction while assuming stable camera trajectory with limited rotational motion. Unfortunately, these assumptions do not hold when a camera is mounted on a UAV, especially when reconstructing challenging environments like bridges and sewage or underground tunnels \cite{samiappan2016using, metni2007uav}.

This paper focuses on reconstructing inaccessible tunnels using a UAV which allows experts to remotely inspect the tunnel's structural integrity and physical conditions. In this application, there are several challenges and limitations that restrict the design of the UAV's imaging setup, and consequently present challenges for traditional SfM algorithms. As we will demonstrate in the results section, SLAM based approach completely fails due to significant rotational motion of the camera. Moreover, SfM based techniques give sub-par results due to having extremely limited amount of images. In fact, we cannot afford to discard a single image as this will result in loss of vital $3$D information of the tunnel. Some proposed solutions \cite{hovermap} require custom built heavy UAVs with multiple light sources, cameras and a LIDAR. This results in a heavy and expensive UAV. Due to safety precautions and limited cross-sectional size of the sewage pipe, we can neither use a LIDAR nor a big and bulky UAV. 

\begin{figure}[t!]
\centering
\includegraphics[width=0.45\textwidth]{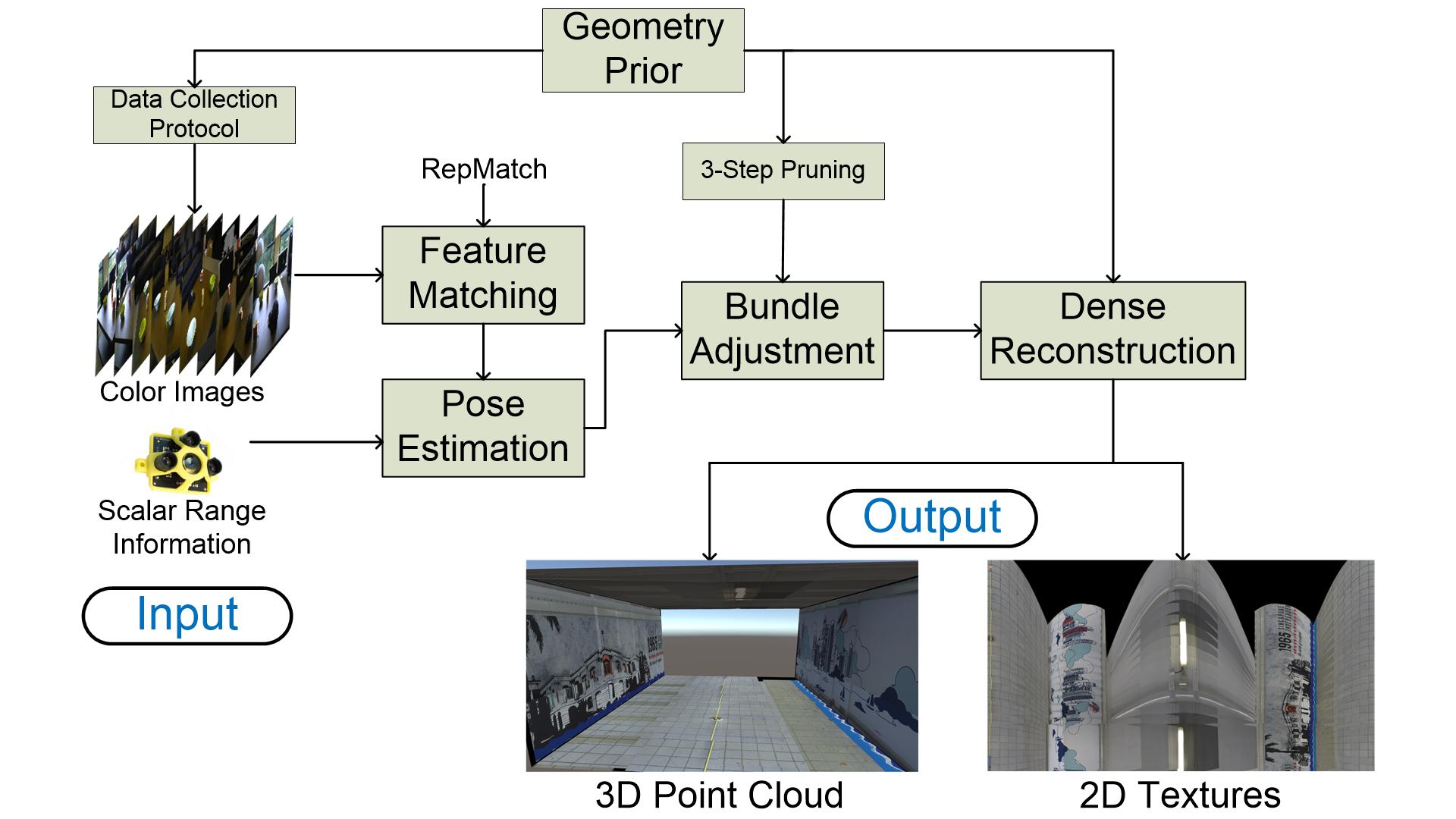}
\caption{Our Dense $3$D reconstruction framework using images collected by a spirally moving camera mounted on a UAV. We exploit geometry prior to assist in data collection, bundle adjustment and $3$D reconstruction process. Our system outputs a $2$D texture map and $3$D pointcloud that can be viewed in Unity for scene visualization and inspection.}
\label{fig::flowchart_ours}
\end{figure}

\begin{figure}[t!]
\centering
\begin{subfigure}[b]{0.15\textwidth}
		\includegraphics[width=\textwidth]{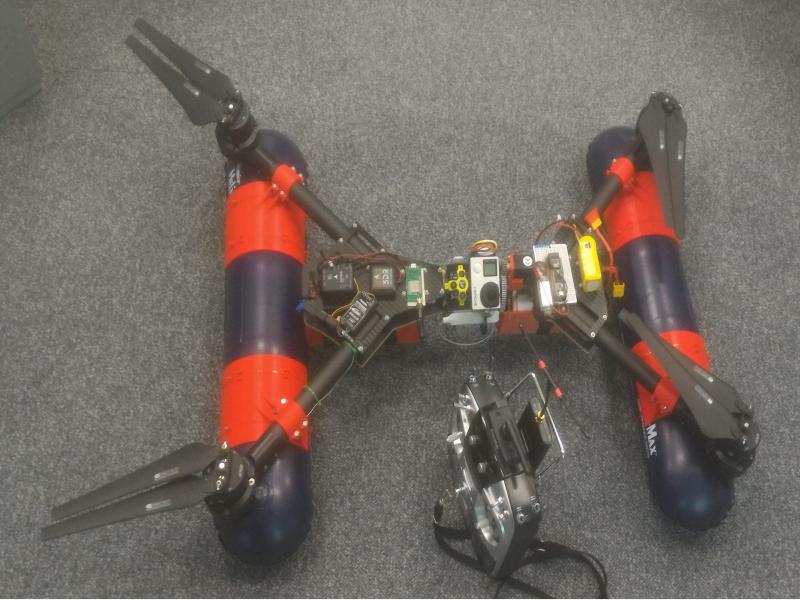}		\caption{Our custom-built UAV.}
	\end{subfigure}   
\begin{subfigure}[b]{0.15\textwidth}
		\includegraphics[width=\textwidth]{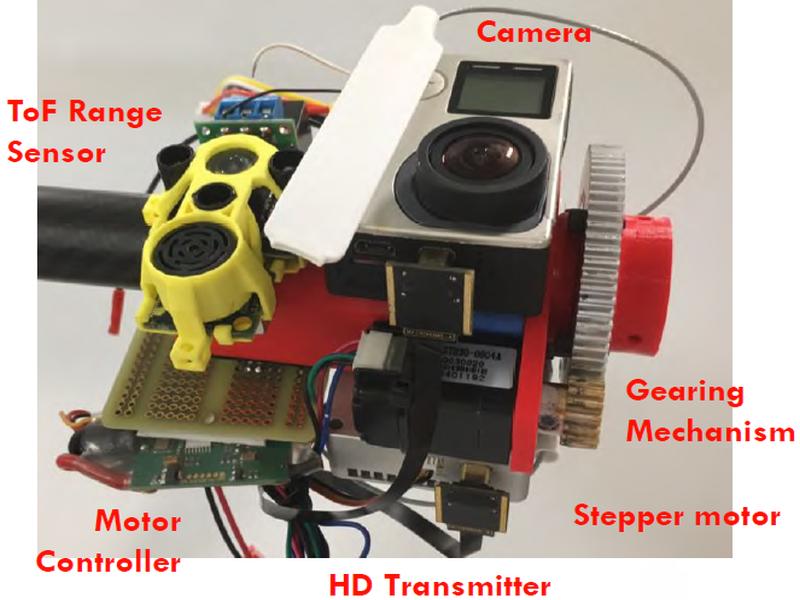} \caption{Rotating mechanism.}
\end{subfigure}	
\begin{subfigure}[b]{0.15\textwidth}
		\includegraphics[width=\textwidth]{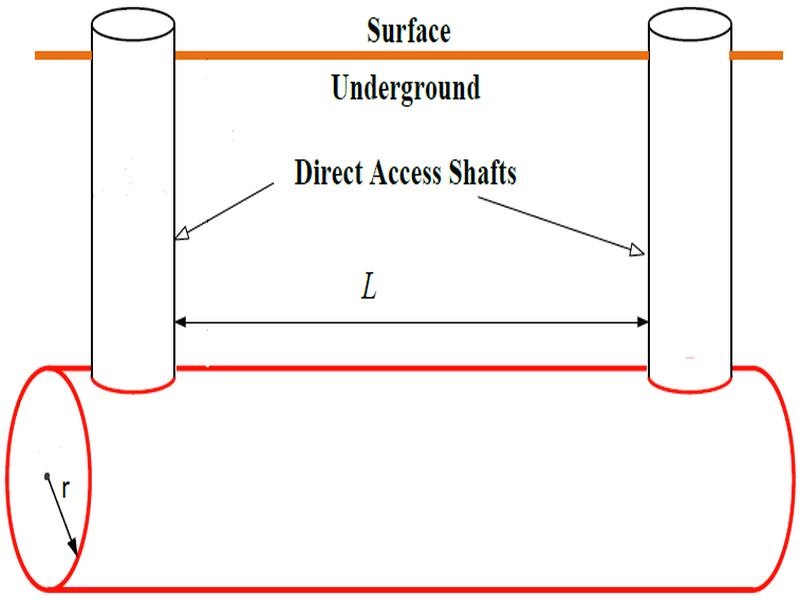} \caption{A sample Cylindrical shaft.}
\end{subfigure}	
\begin{subfigure}[b]{0.45\textwidth}
		\includegraphics[width=\textwidth]{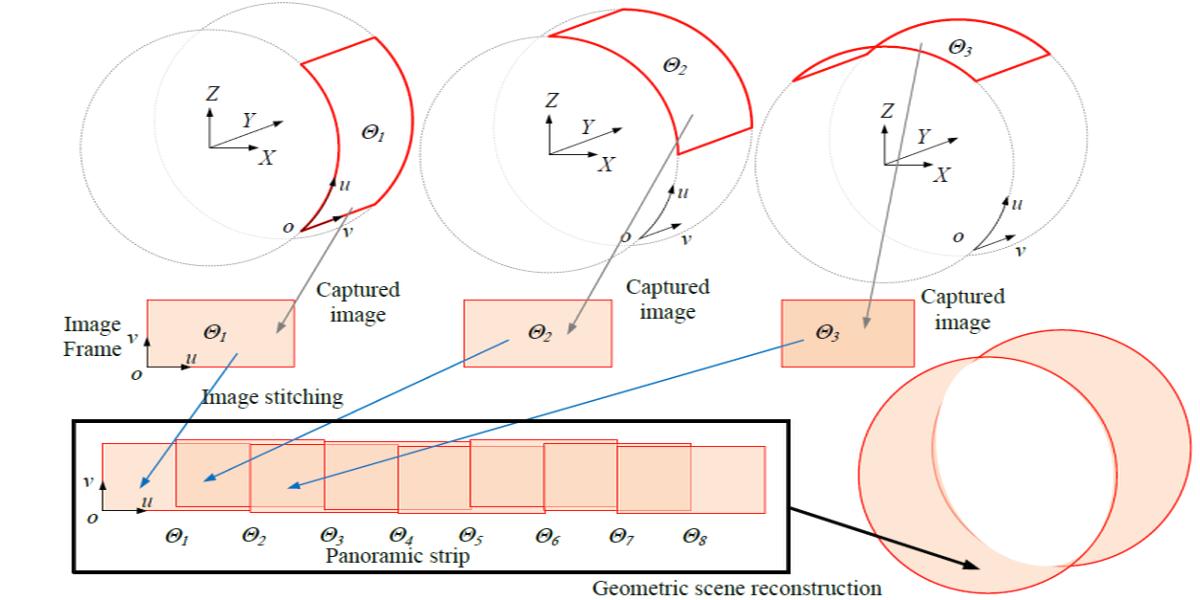} \caption{Cylindrical dataset.}
\end{subfigure} 
\begin{subfigure}[b]{0.45\textwidth}
		\includegraphics[width=\textwidth]{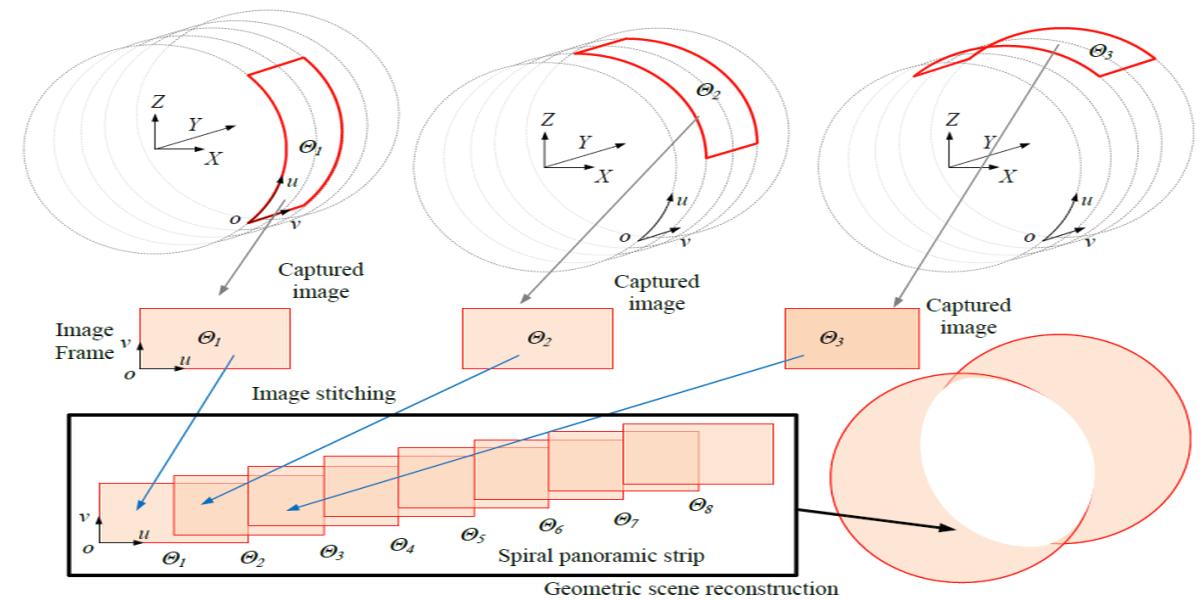} \caption{Spherical dataset.}
\end{subfigure}
\caption{Our UAV consists of a rotating shaft with a GoPro camera rigidly connected to a range sensor and light source. The shaft rotates the camera and range sensor resulting in a ``Cylindrical dataset" if the UAV is static; and a ``Spiral dataset" if the UAV flies horizontally in a given environment resulting in a panoramic spiral image.} 
\label{fig::UAV_design_framework}
\end{figure}

Thus, taking into consideration limitations on power, weight, mandatory safety features and resolution, we custom-built a UAV designed for long flight time and use a light-weight GoPro camera \cite{gopro4_url}. We rotate the GoPro around the shaft of the UAV while the UAV traverses the tunnel. This paper presents a framework, as shown in Fig.~\ref{fig::flowchart_ours}, to densely reconstruct a given scene in $3$D using our custom UAV. The input to our system would be a series of images taken with an outward-facing camera moving along a spiral trajectory as shown in Fig.~\ref{fig::UAV_design_framework}.
We utilize Structure-from-Motion (SfM) along with Bundle Adjustment (BA) and a geometry prior to reconstruct the scene geometry and generate cylindrical images for texture mapping. The geometry and textures are used in our $3$D visualization tool in Unity \cite{Unity2017} to aid experts in visualization, remote inspection, and fault detection of these tunnels.

In particular, we make the following contributions:
\begin{itemize} 
\item A framework to estimate maximum theoretical speed for complete coverage and method to centralize the UAV in the tunnel using range sensors.
\item Utilize geometry prior during BA to improve camera pose estimation.
\item Dense $3$D Reconstruction of the scene using geometry prior where all pixels are estimated.
\end{itemize} 
We evaluate our approach on both synthetic dataset where ground-truth information is available, and real data obtained using a rotating camera setup and compare with state-of-the-art reconstruction methods. In both cases, we demonstrate that our proposed framework is capable of handling actual situations where the camera data contain noise and jitters. 


\begin{section}{Related Work}\label{sec::related_work}
\noindent \textbf{SfM and SLAM: } Structure-from-Motion (SfM) refers to the recovery of $3$D structure of the scene from given images. A widely known application of SfM is where ancient Rome is reconstructed using Internet images \cite{agarwal2011building, frahm2010building, kushal2012photo, wu2013towards}. SfM is highly dependent on accurate feature matching to estimate the camera's pose and location. RANSAC \cite{fischler1981random} is typically used for outlier detection which unfortunately results in discarding a large fraction of true matches \cite{lin2014bilateral}. This results in a fragmented and discontinuous reconstruction \cite{wu2011visualSfM}. 

Another approach to estimate camera pose is by using Simultaneous Localization and Mapping (SLAM) \cite{orb_slam_paper, dense_reconstruction_fly, kerl2013dense}. SLAM usually works in situations where the camera captures a video sequence consisting of minimal rotational motion. Unfortunately, both these assumptions do not hold true for our case. Off-the-shelf SLAM algorithms such as ORB-SLAM \cite{orb_slam_paper} often fail to estimate camera pose for our synthetic and real data experiments. Our solution is to use \textit{RepMatch} \cite{lin2016RepMatch} - a robust wide-baseline feature matcher which uses an epipolar guided feature matcher to guide the discovery of more feature matches. Repmatch works exceptionally well in man-made repeated structures and provides a dense set of matching feature points between a pair of images. This allows us to reliably find dense correspondence between images consisting of significant camera rotation between them.
\newline
\newline
\noindent \textbf{Bundle Adjustment}: In general, Bundle Adjustment (BA) refers to the problem of jointly refining and estimating optimal $3$D structure and camera(s) intrinsic and extrinsic parameters. Classically, BA is formulated as a non-linear least squares problem \cite{ engels2006bundle}. The cost function is assumed to be quadratic in terms of $2$D reprojection error. Outlier detection and removal are used to make it more robust to noise and missing data. We leverage on the scene geometry to make our system more robust and slightly faster than general BA. During the triangulation process, we aggressively discard points with a significant $2$D reprojection error. We also detect and discard points that are farther away than the expected geometry. This not only makes the BA results more accurate, but also results in a faster convergence as there are considerably fewer $3$D points to fit in the non-linear least squares problem.
\newline
\newline
\noindent \textbf{3D Reconstruction: } Once the camera pose is robustly estimated, all the matching feature points are triangulated in $3$D. This involves triangulating multiple views of the same region for a semi-dense reconstruction of the scene. Some approaches  \cite{kolev2012fast} work with volumetric representations and usually do not scale very well to large scenes while others such as Multi-View Reconstruction Environment (MVE) break down the scene into multiple segments \cite{furukawa2010towards, fuhrmann2014mve}. MVE integrates SfM, multi-view stereo (MVS), and floating scale surface reconstruction (FSSR) to produce a surface triangle mesh as output. First, SfM is used to reconstruct camera parameters and a sparse set of matching points. Depth maps are computed using MVS and a colored mesh is extracted from depth maps using FSSR resulting in a semi-dense $3$D reconstruction. All points are kept in memory in order to be used in evaluation of the implicit function. This memory overhead prevents MVE from handling datasets which contain more than a few hundred images. Most of these approaches are based on low-level vision and hence do not understand the scene. Our approach adds another component by exploiting the scene geometry. All points, corresponding to every pixel lying an image, are estimated to fit the geometry and used to reconstruct the scene.  

\end{section}

\begin{section}{Data Acquisition} \label{sec::cyl_math}
The quality of the texture map generated is paramount for Virtual Reality applications. This means that the final texture map must have high resolution and complete coverage. In the case of a UAV acquiring the input data for reconstructing a tunnel, the position of the UAV must be at the center of the tunnel cross-section to have maximum resolution for all views. The movement of the UAV is also crucial because there must always be overlapping regions between when the camera completes a full rotation. 

\begin{subsection}{UAV camera set-up}\label{subsec::deep_tunnel_imaging}
While a detailed analysis of our custom UAV design is out of scope, we briefly describe major constrictions that informed the current UAV design. The imaging system has to be lightweight, high-resolution and have an unobstructed $360\degree$ view. Commercial UAV cameras are not suitable because they are designed for high-altitude long-range imaging and limited field-of-view. Augmenting the UAV with high resolution commercial $360\degree$ cameras is an alternative but they are too bulky, consume too much power, and are difficult to mount onto the UAV. In fact, since our application requires maximum resolution along the walls of the tunnel, significant percentage of pixels from $360\degree$ cameras would not be usable, further reducing the final resolution.

Our solution mounts a camera perpendicular to the intestine surface (UAV shaft) where there is no obstructed view and image distortion is minimal at areas of interest. A light source is rigidly attached to the camera that illuminates the camera's FoV. The camera rotates and captures image data while the UAV moves forward (Fig.~\ref{fig::UAV_design_framework}(c)) to capture the entire inner cylindrical surface of the tunnel with high fidelity and minimal optical distortion. We also tried using four synchronized cameras with four corresponding light sources instead of a rotating shaft. However the extra weight of cameras and light sources along with reduced battery life make it a non-viable option. In practice, we use a GoPro HERO 4 camera that is programmed to capture an image after rotating a certain degree every few milliseconds. This set-up provides us with spiral-like images and produces $7,500$ pixels per $360\degree$ resolution after stitching. It translates to an angular resolution of $0.0325\degree$ per pixel and approximately $1.70$mm per pixel resolution for the lateral movement in a tunnel of radius $3$m. 
\end{subsection}
\begin{subsection}{Geometry Prior} \label{sec::geometry_prior}
One of the key components of our work is the use of geometry prior at various stages of the proposed framework as shown in Fig.~\ref{fig::flowchart_ours}. The prior known geometry such as tunnel radius and wall dimensions of a rectangular room enables us in achieving various components that would not be possible otherwise. In particular:
\begin{enumerate}
    \item Geometry prior enables us to perform dense $3$D reconstruction of the entire scene. As shown in \cite{raman_icis}, known geometry prior can also allow for feature-less stitching of $2$D images.  
    \item It allows us to pre-configure our UAV's speed for most efficient data capture as explained in Sec.~\ref{sec::max_uav_movement}. 
    \item It also assists us in removing erroneous triangulated points during the Bundle Adjustment process. Any $3$D triangulated point that is far away from expected geometry such as tunnel boundary is removed from the Bundle Adjustment process as discussed in Sec.~\ref{sec::BA}.
\end{enumerate}
\end{subsection}

\begin{subsection}{Sectional Location of the UAV in the Tunnel}
\begin{figure}[t!]
\centering
\begin{subfigure}[b]{0.225\textwidth}
\includegraphics[width=\textwidth]{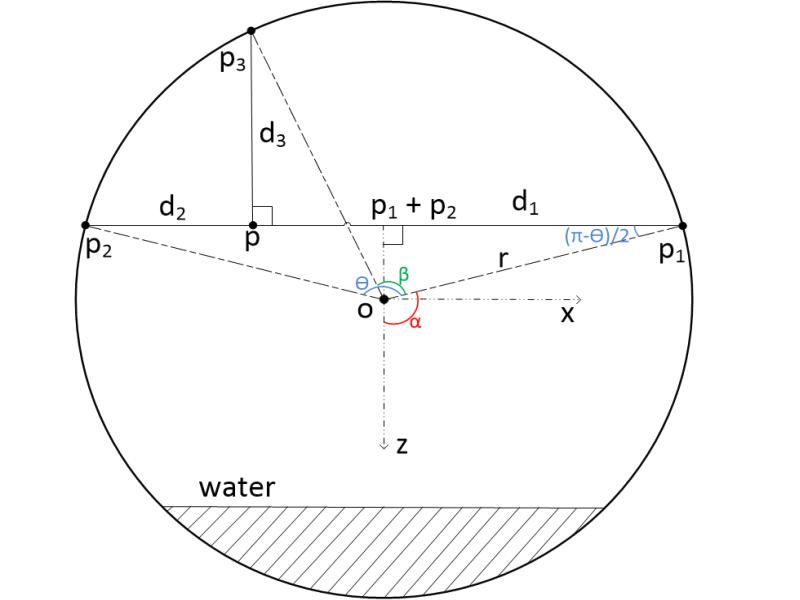}
\caption{}
\end{subfigure}
\begin{subfigure}[b]{0.225\textwidth}
\includegraphics[width=\textwidth]{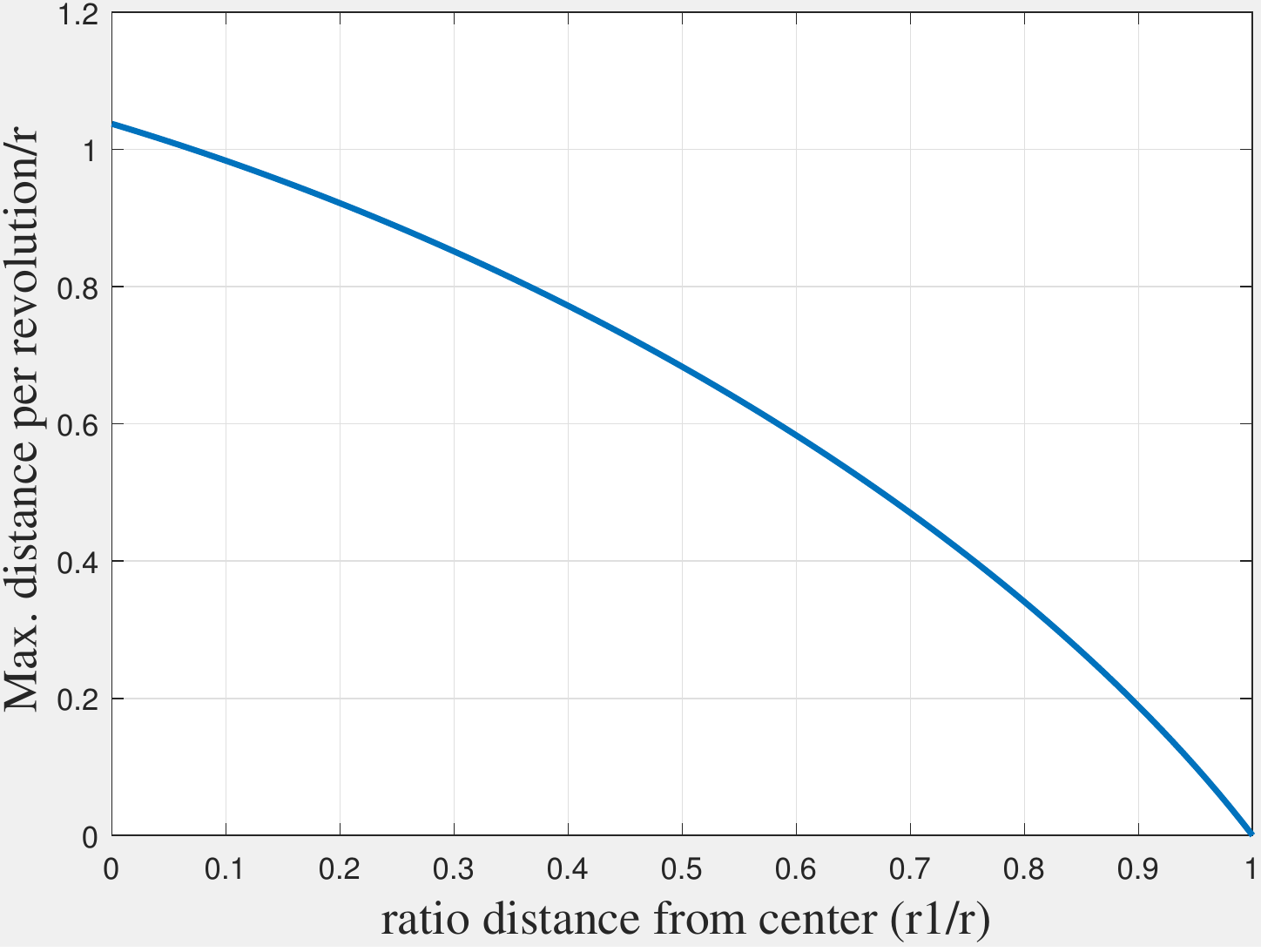}
\caption{}
\end{subfigure}
\caption{We showcase our framework to initialize the position and estimate the maximum speed of our UAV. (a) The initial position of the UAV is triangulated using the depth sensors located on the $4$ sides of the UAV. (b) As the distance, $r_1$, from center of tunnel increases, the maximum allowable distance moved by the UAV decreases to ensure no dead-space during the offline stitching process.}
\label{fig::UAV_framework} 
\end{figure}
We use four range sensor on the UAV to place it in the center of the tunnel's cross-section. These range sensors measure the distance of the UAV vertically and horizontally (Fig.~\ref{fig::UAV_framework}(a)) and have a range up to $14$m. We use the in-built inertial measurement unit (IMU) to stabilize the UAV and point it horizontally forward, i.e. zero pitch, yaw, and roll. We discard the readings from the range sensor pointing downwards and use the range information from remaining three sensors since there may be flowing water or sludge in the tunnel.

Assume the $3$D location of the UAV to be $\bm{p}$, the distance measured by two horizontal and a vertical range sensors be $\{d_1, d_2, d_3\}$ respectively, and the radius of the tunnel be $r$. We present a framework to compute the offset of the UAV from tunnel center, $[t_x,t_z]^\intercal$. We parameterize the three points located by the three range sensors as follows:

\begin{align}
{\scriptscriptstyle \bm{p}_1 =   \begin{bmatrix} r \sin\alpha \\ h \\ r \cos \alpha \end{bmatrix}, \quad  \bm{p}_2 =  \begin{bmatrix} r \sin(\alpha+\theta)  \\ h \\ r \cos(\alpha+\theta) \end{bmatrix}, \quad \bm{p}_3 =   \begin{bmatrix} r \sin(\alpha+\beta)  \\ h \\ r \cos(\alpha+\beta) \end{bmatrix}; }  
\end{align}

As the UAV lies on $(\bm{\vec{p}}_2-\bm{\vec{p}}_1)$, the $3$D point, $\bm{p}$, is:
\begin{align}\small
\bm{p} &= \begin{bmatrix} t_x \\ 0 \\ t_z \end{bmatrix} = \gamma \bm{p}_1 + (1-\gamma) \bm{p}_2 \nonumber \\ 
 &= \begin{bmatrix} r (\gamma \sin\alpha + (1-\gamma)\sin(\alpha+\theta))  \\ 0 \\ r (\gamma \cos \alpha +(1-\gamma) \cos(\alpha+\theta)) \end{bmatrix}; \gamma \in [0,1] 
\end{align}
where $\{\theta,\alpha, \beta, \gamma \}$ are four unknown variables. Since the UAV is stabilized with zero yaw, pitch and roll, $h$ is assumed to be zero. While a generic circle can be defined by three given points, we only need two points to locate the UAV's position in the tunnel because we know the tunnel's coarse dimensions. We use the range sensor information along with the tunnel radius as follows:
\begin{align}
\begin{bmatrix} ||\bm{p}-\bm{p}_1||_2 - d_1 \\
||\bm{p}-\bm{p}_2||_2 -  d_2 \\
||\bm{p}-\bm{p}_3||_2 -  d_3 \\
\cos\theta - \frac{r^2+r^2-(d_1+d_2)^2}{2\cdot r \cdot r}  \\ 
\frac{\theta}{2} +\alpha-\pi  \\
(\bm{p}-\bm{p}_3) \times (\bm{p}_1+\bm{p}_2)  \end{bmatrix} = \vec{0}
\end{align}
This results in an over-constrained solution for $\{\theta, \alpha, \beta,\gamma\}$. We use ordinary least squares (OLS) \cite{lawson1995solving} to estimate the parameters. This approach is robust to sensor failures as the system is over-constrained, potentially saving resources when deploying it.
\end{subsection}

\begin{subsection}{Maximum Speed of UAV} \label{sec::max_uav_movement}
If the UAV moves too fast while the camera is rotating, there will be slivers of incomplete coverage where the camera did not capture any data. In this section, we present our framework to estimate the maximum distance the UAV is allowed to move to avoid this undesirable outcome. 

First, we estimate the maximum distance $d$ that the UAV can move between the two ``consecutive" images after a complete rotation (Fig.~\ref{fig::UAV_framework}(b)). The two captured images must contain some overlap so they can be reconstructed without having any dead space (black holes) between them. We extrapolate the $3$D points represented by the extreme pixel locations - $[0,0]^\intercal$ and  $[0,nr)^\intercal$ - as follows:
\begin{align}
\begin{bmatrix} 0 \\ 0 \end{bmatrix} = \begin{bmatrix} f\frac{X_1}{Z_1}+c_x \\ f\frac{Y_1}{Z_1}+c_y \end{bmatrix} ; \begin{bmatrix} 0 \\ nr-1 \end{bmatrix} = \begin{bmatrix} f\frac{X_{nr}}{Z_{nr}}+c_x \\ f\frac{Y_{nr}}{Z_{nr}}+c_y \end{bmatrix}
\end{align}
where $[X_1,Y_1,Z_1]^\intercal$ and $[X_{nr},Y_{nr},Z_{nr}]^\intercal$ represent the $3$D points located on pixels - $[0,0]^\intercal$ and $[0,nr)^\intercal$ respectively, $f$ represents the focal length of the camera and $[c_x,c_y]$ represents the optical center of the camera. Since we want at least some overlap between two consecutive images, the maximum distance that the UAV can move is represented by:
\begin{align}
d_{max} = \Delta Y = Y_{nr}-Y_1 = \frac{(nr-1)(r\cos \theta - r_1)}{f} 
\end{align}
where $r_1$ refers to the current distance between the UAV and the tunnel center. Using law of sines as seen in Fig.~\ref{fig::UAV_framework}(b), we can estimate $\theta$ as follows:
\begin{align}
\theta = \frac{\Omega_h}{2} -\arcsin \left(\frac{r_1}{r}\sin(\frac{\Omega_h}{2}) \right)
\end{align}
where $\Omega_{h}$ refers to the horizontal field of view of the camera, $r_1$ refers to distance of the UAV from the center of the tunnel, and $r$ represents the tunnel radius. Thus, the maximum horizontal movement allowed per $360\degree$ rotation is:
%
\begin{align}
d_{max} \approx 2 \tan(\frac{\Omega_{v}}{2}) \left(r\cos\theta-r_1 \right)
\end{align}
%
\begin{figure}[t!]
\centering  
\begin{subfigure}[b]{0.225\textwidth}
		\includegraphics[width=\textwidth]{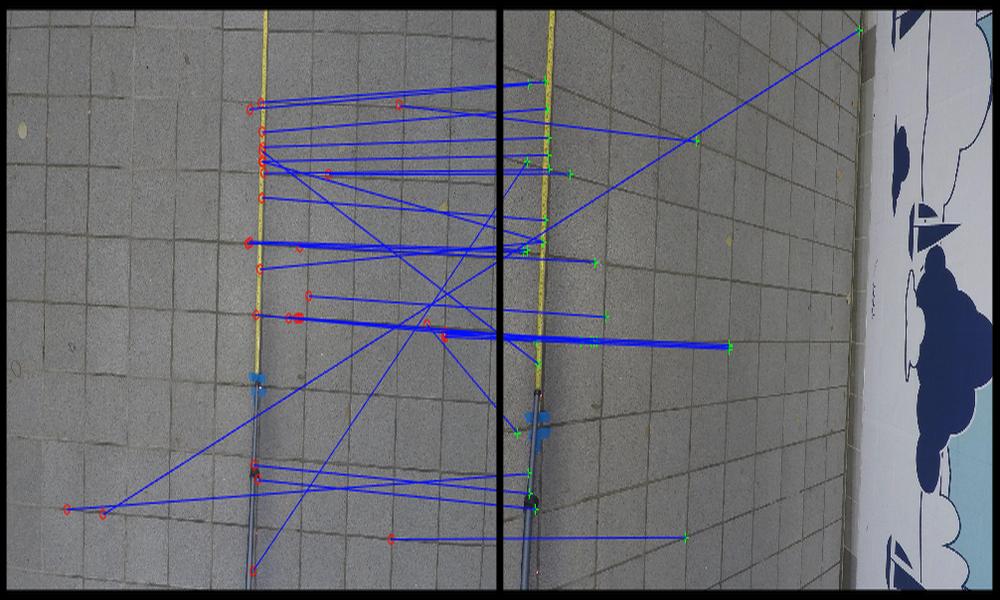}
		\caption{SURF matching points}
	\end{subfigure}   
	\begin{subfigure}[b]{0.225\textwidth}
		\includegraphics[width=\textwidth]{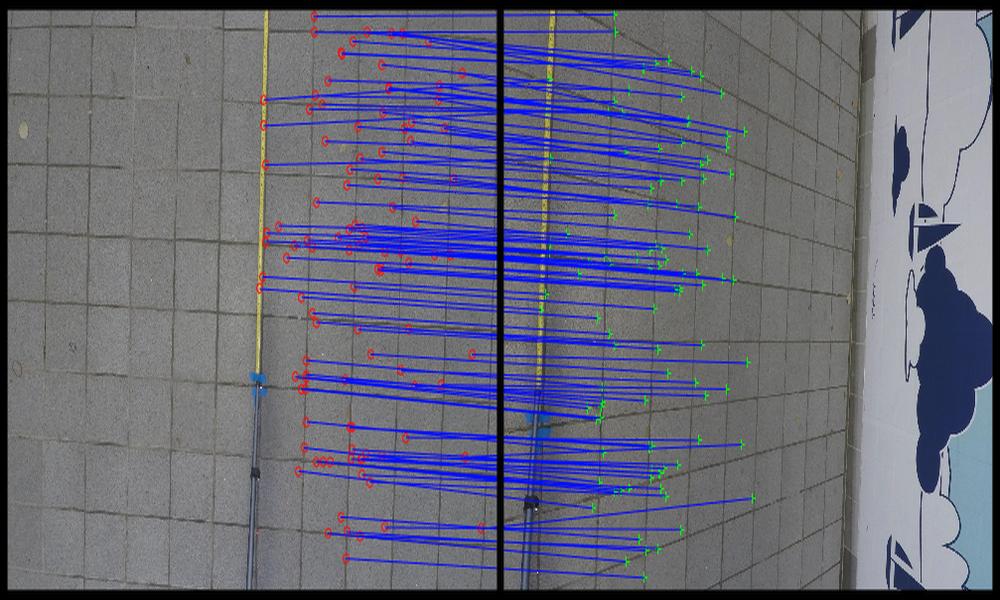}
		\caption{RepMatch matching points}
	\end{subfigure}   
  \caption{We obtained $31$ matching points using SURF, while RepMatch resulted in $10,651$ matching points. We only display $1\%$ RepMatch points for visualization. We observe that SURF based matching struggles in presence of repeated structures (floor tiles) while RepMatch gives us a more accurate and dense point matching. }
 \label{fig::surf_vs_repmatch}
\end{figure}
%
\newline \noindent where $\Omega_{v}$ refers to the vertical field of view of the camera. Given $n$ images per full $360\degree$ rotation of the camera, the maximum movement allowed per image is:
\begin{align} \label{eq::max_movement}
\hat{d}_{max} \approx\frac {2\cdot (r\cos \theta-r_1) \tan(\frac{\Omega_v}{2})}{n}
\end{align}
Fig.~\ref{fig::UAV_framework}(c) shows how this maximum distance varies as we go farther from the tunnel center. Intuitively, the farther we are from center, the slower UAV needs to be as the UAV is closer to some sections of the tunnel and hence may not be able to see the entire region if the UAV is moving too fast.
\end{subsection}

\end{section}

\begin{section}{3D Reconstruction} \label{sec::our_algorithm}
Our $3$D reconstruction algorithm improves camera pose estimation for difficult scenes by using a geometry prior along with other improvements during the bundle adjustment process. We also present a texture mapping approach which avoids memory overhead when processing and storing $3$D pointclouds.

\begin{subsection}{Camera Pose Estimation}
Camera pose refers to change in orientation and position of the moving camera between two adjacent frames. It can be estimated using SfM techniques \cite{agarwal2011building, frahm2010building, kushal2012photo, wu2013towards}. First, local features such as SIFT \cite{Lowe1999}, SURF \cite{bay2008speeded} or ORB \cite{ORB_features} are computed for each image. Subsequently, each feature is matched with the best corresponding feature in the adjacent image. Matches that are poor and fail a distance ratio test (greater than $0.8$) are rejected as false matches. Finally, RANSAC is used to identify and reject outliers, and the camera pose is estimated using the remaining inliers. 

Unfortunately, current feature matching techniques struggle with estimating accurate camera poses in presence of repeated structures such as floor tiles or brick walls. This is because repetitive man-made structures, such as brick walls or tiles, appear similar resulting in an inaccurate pose estimation.
Thus, we use RepMatch \cite{lin2016RepMatch} to estimate camera pose as it is shown to perform better in such situations. RepMatch couples Bilateral Functions (BF) \cite{lin2014bilateral} and RANSAC outlier rejection schemes by relaxing the ratio test. A comparison of SURF and RepMatch feature point matching is shown in Fig.~\ref{fig::surf_vs_repmatch}.
\end{subsection}

\begin{figure}[t!]
\centering
\begin{subfigure}[b]{0.225\textwidth}
		\includegraphics[width=\textwidth]{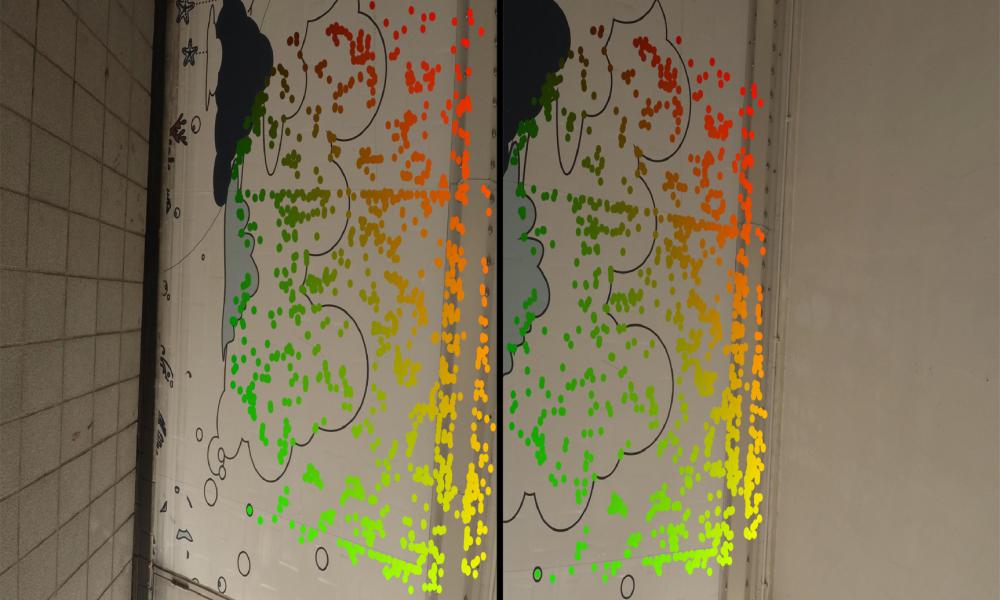}
		\caption{$36 \degree$ rotational motion.}
	\end{subfigure}    
\begin{subfigure}[b]{0.225\textwidth}
		\includegraphics[width=\textwidth]{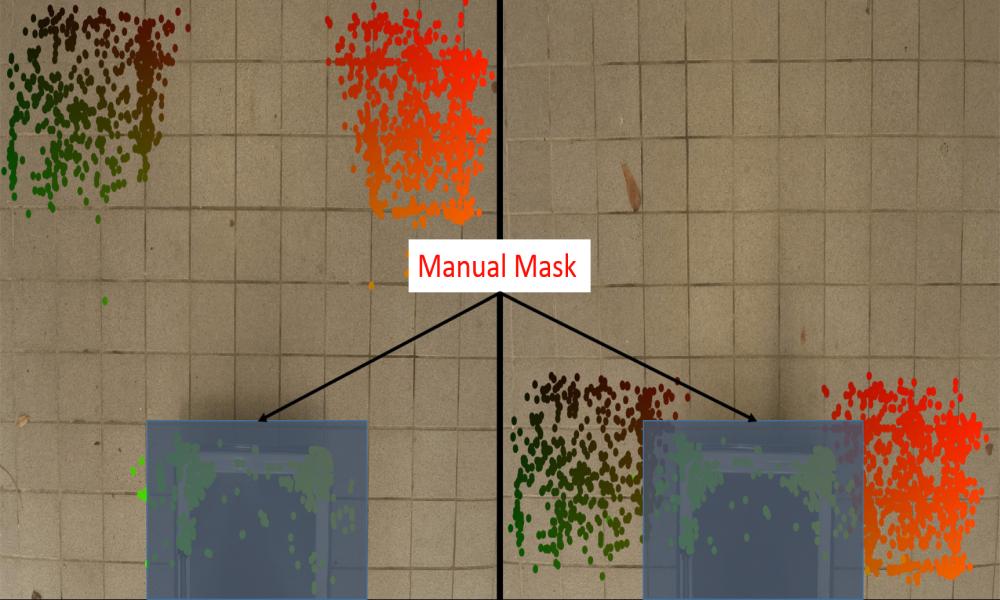}
		\caption{Translational motion.}
	\end{subfigure}  	
  \caption{(a) RepMatch correctly identifies matching points resulting in a good camera pose estimate. (b) In scenes where objects, such as trolley, are static with respect to the camera, it incorrectly identifies them as matching points resulting in a noisy camera pose. We avoid these wrong matches by using a manual mask. }
\label{fig::RepMatch}
\end{figure}
\begin{subsection}{Bundle Adjustment} \label{sec::BA}
Bundle Adjustment (BA) triangulates the matched points to estimate their $3$D location by minimizing the $2$D projection error across all images containing them. Since a maximum of four images share a partial view with any given image, it is critical to avoid any outliers as BA could fail due to erroneous or noisy matching. To improve robustness in our system, we perform a three-step pruning process: 
\begin{itemize}
\item Pruning1: Mask out relatively static objects such as UAV components in camera's FOV. This is done by utilizing prior knowledge of static objects locations and camera's FOV. 
\item Pruning2: Prune erroneous triangulated points that violate the geometry prior. Any points that are triangulated to be far away from expected geometry such as tunnel walls are rejected.
\item Pruning3: Reject matching points with large reprojection error as we are only interested in camera pose rather than scene reconstruction. 
\end{itemize}
For example, in Fig.~\ref{fig::RepMatch} we mask out the ROI occupied by trolley as it is static with respect to the moving camera (Pruning1). Any triangulated points that are far off from the estimated underpass region are rejected as well (Pruning2). Moreover, any points with significantly high reprojection error are also thrown out (Pruning3). This three-step pruning results in a significantly lower average reprojection error as seen qualitatively in Fig.~\ref{fig::floor_BA} and quantitatively in Table~\ref{table::reprojection_error} on our datasets described in Sec.~\ref{sec::real_datasets}. For a fair comparison, we do not show the results for Pruning3 since it removes the matching points with high reprojection error after BA. 

Standard Bundle Adjustment (SBA) works well in cylindrical dataset because the camera is stationary for one complete rotation and only moved thereafter. SBA has no issues with this setup because $90\%$ of the frames have no translation. However, in spiral dataset, the camera is rotated significantly and translated for each image. The SBA process becomes brittle because a few outliers can induce major errors. Therefore, it is critical to remove these outliers during the BA process. In the pruning, reconstructed points that triangulate far away from geometry are removed. This improves our results significantly.
\begin{figure}[t!]
\centering  
\includegraphics[width=0.45\textwidth]{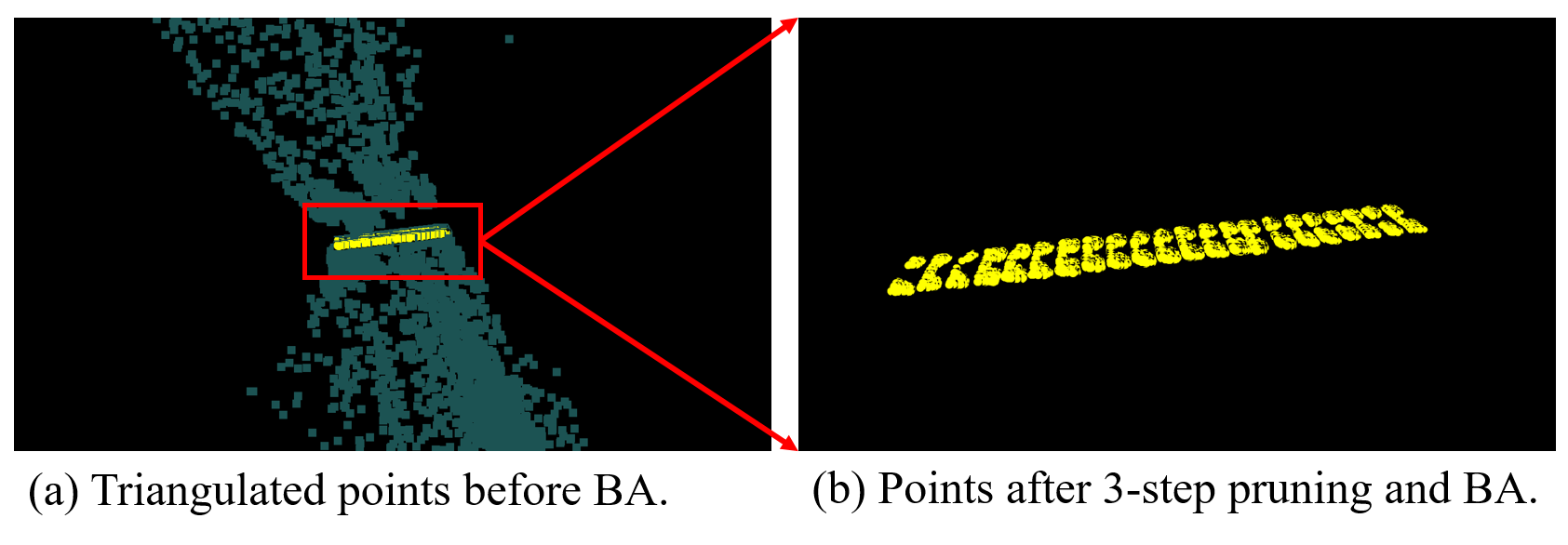}
  \caption{ (a) $3$D triangulated points lying on floor in the trolley dataset before SBA (shown in green). (b) $3$D triangulated points after three-step pruning and BA. Points not lying close enough to the floor (in blue) are removed. Our three-step pruning removes approximately $24,000$ inaccurate points out of $105,000$ points in the trolley dataset. }
\label{fig::floor_BA}
\end{figure}
\begin{table}[htbp]
\centering
{ \begin{tabular}{ |c||c|c|c|c|} \hline 
{Dataset:} &  \multicolumn{4}{|c|}{Cylindrical}  \\ \hline 
{} & SBA & $P1$ & $P2$ &  $P1+P2$  \\ \hline 
Before BA  & $22.83$ & $34.33$ & $36.18$ & $30.81$  \\ \hline 
After BA  & $0.74$ & $0.74$ & $0.65$& $\bm{0.63}$   \\ \hline 
{Dataset:}  & \multicolumn{4}{|c|}{Spherical}  \\ \hline
Before BA  & $46.54$ & $62.63$ & $51.46$ & $79.68$ \\ \hline
After BA & $41.88$ & $\bm{0.69}$ & $1.65$& $3.25$ \\ \hline
\end{tabular}
}
\caption{Ablation study of our Pruning process for BA. The $2$D reprojection error before and after BA on our datasets is described in Sec.~\ref{sec::real_datasets}. $P1$ refers to Pruning1, $P2$ refers to Pruning2, and $P1+P2$ refers to performing both Pruning1 and Pruning2 steps together. Our framework reduces the error significantly when compared to Standard Bundle Adjustment (SBA).
}
\label{table::reprojection_error}
\end{table}
%
\vspace{-0.5em}
\end{subsection}

\begin{subsection}{Texture mapping}
We can further exploit the geometry prior to save memory and storage overhead. Instead of saving the fully-dense $3$D pointcloud data, we compute a cylindrical projection $2$D image for tunnels and multiple planes for indoor rooms and underpasses \cite{raman_icis}. These images are used in the Unity engine \cite{Unity2017} as texture maps to recreate the scene which allows the user to navigate, zoom in and out of the scene for visual inspection and anomaly detection \cite{raman_icis}.   
\end{subsection}

\end{section}

\begin{section}{Experimental Results}\label{sec::exper_results}
We perform both synthetic experiments, as a proof of concept, and real-data experiments to demonstrate the robustness of our approach in various scenarios. We also compare our results with state-of-the-art SfM method - MVE \cite{fuhrmann2014mve} - using both synthetic and real data captured in an underpasses that highlight the accuracy and robustness of our approach. Using simulations allows us to capture groundtruth camera trajectory along with 3D geometry of the tunnel to verify the 3D reconstruction process and to determine the UAV's maximum speed.  We display a few examples of the rendered scenes in Unity for visualization.  

The SLAM algorithms fail to estimate camera pose due to significant rotational motion. Based on our experiments in Unity, ORB-SLAM allows a maximum of $2\degree$ rotational motion, in a $3$m radius tunnel, per image for accurate camera trajectory estimation. In comparison, we rotate our camera approx. $36\degree$ per image. Hence, we do not show a visual comparison with SLAM based reconstruction. 

\begin{subsection}{Synthetic Data Results}
We used blender\cite{blender_cite} to render a hollow cylinder and imported it into Unity \cite{Unity2017} to generate our synthetic data. We texture-mapped a brick wall and a panoramic view of Seattle's skyline onto the inner face of the cylinder for visualization purposes. The light source is fixed to look vertically downwards, resulting in images captured with $0\degree$ rotation to be brightly lit while the images captured with $180\degree$ rotation to be dark. 
\begin{subsubsection}  {Varying the Speed of UAV} 
In this experiment, we vary the speed of the UAV to empirically show that Eq.~\ref{eq::max_movement} provides us with a reasonable bound for maximum UAV speed to obtain a full panoramic stitch of the tunnel without any holes. We simulated a camera with focal length $3800$ pixels, FoV of $[\Omega_h,\Omega_v] = [60\degree, 101\degree]$, in a tunnel of radius $3$m. Using Eq.~\ref{eq::max_movement}, we obtain a maximum allowable speed of approximately $18$cm per rotation for the UAV. We simulated the camera to move horizontally forward at $18$cm per rotation and $20$cm per rotation. We only cylindrically project \cite{raman_icis} the $0\degree$ and $180\degree$ images to illustrate the impact of UAV's speed on coverage in Fig.~\ref{fig::speed_stitch}. We obtain a seamless stitch without any holes when the camera moves $18$cm per rotation. However, as soon as we increase the speed to $20$ cm per rotation, we start seeing gaps between the projected images as the camera is moving too fast and certain sections of the tunnel are left unseen.
\end{subsubsection}
\begin{figure}[t!]
\centering
	\begin{subfigure}[b]{0.225\textwidth}
		\includegraphics[width=\textwidth]{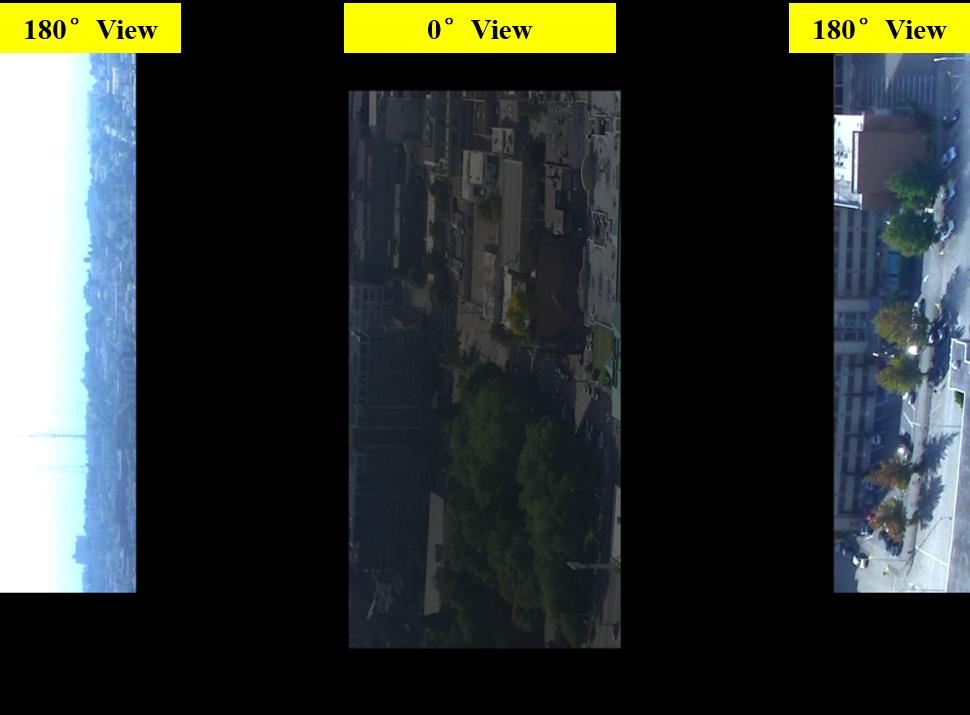}
		\caption{}
	\end{subfigure}  
	\begin{subfigure}[b]{0.225\textwidth}
		\includegraphics[width=\textwidth]{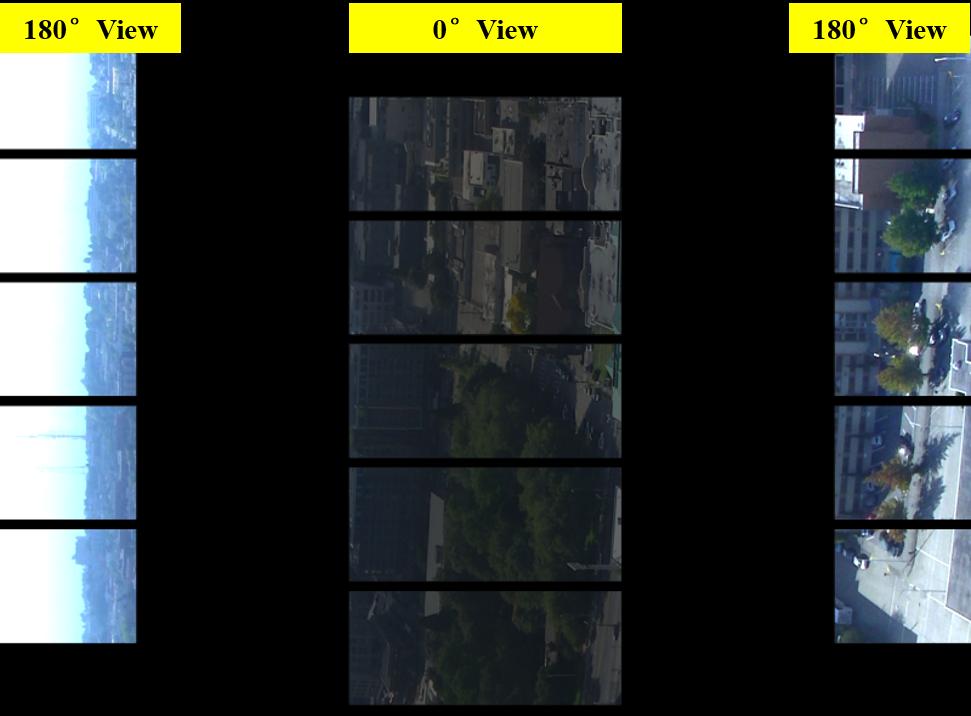}
  \caption{}
  	\end{subfigure}  
\caption{Cylindrical projection when the camera moves forward by (a) $18$cm (b)$20$cm per rotation.  We only stitch the $0\degree$ and $180\degree$ images to highlight the gaps between views. As the speed of the camera increases above maximum estimated speed, we start observing holes in the cylindrical projection.}
\label{fig::speed_stitch}
\end{figure}
\begin{subsubsection}{Comparison with MVE} 
In this experiment, we aim to simulate UAV movements in real-world conditions and compare our framework with MVE. A UAV is expected to suffer from jitters and sideways movements while it tries to balance itself and move forward in the tunnel. 

The camera is positioned at the center of the simulated cylindrical tunnel of $3$m radius. Our baseline movement and rotation in the three orthogonal directions are $t = [0,0.15,0]^\intercal$ and $r =[0,0.524,0]^\intercal$ respectively. With this set-up, the ideal movement of the UAV is $15$cm horizontally forward (y direction) with a $30\degree$ rotation across y axis. We add Gaussian noise with zero mean and variance of $[2,1,2]^\intercal$ to our translation and Gaussian noise with zero mean and variance of $2\degree$ to our rotation per image. We run the simulation for ten camera rotations and record the groundtruth translation and rotation of the camera per frame as shown in Fig.~\ref{fig::SfM_stitch_seattle}(a-b). Two cylindrical projections are computed from the data using MVE, and RepMatch.
 
The $3$D pointcloud from MVE is used for cylindrical projection to visualize the results in $2$D as shown in Fig.~\ref{fig::SfM_stitch_seattle}(c). In our second approach, we use RepMatch to perform wide-baseline feature point matching across two consecutive images to estimate the camera pose. This camera pose is further used for 3D reconstruction and cylindrical projection as shown in Fig.~\ref{fig::SfM_stitch_seattle}(d). While MVE pose estimation does a good job, it outputs a semi-dense reconstruction due to various areas lacking distinct feature points. RepMatch also outputs an accurate camera pose resulting in a fully-dense $3$D reconstruction which is essential for tasks such as visual inspection and anomaly detection. 
\end{subsubsection}
\begin{figure}[t!]
\centering
   \begin{subfigure}[b]{0.225\textwidth}
		\includegraphics[width=\textwidth]{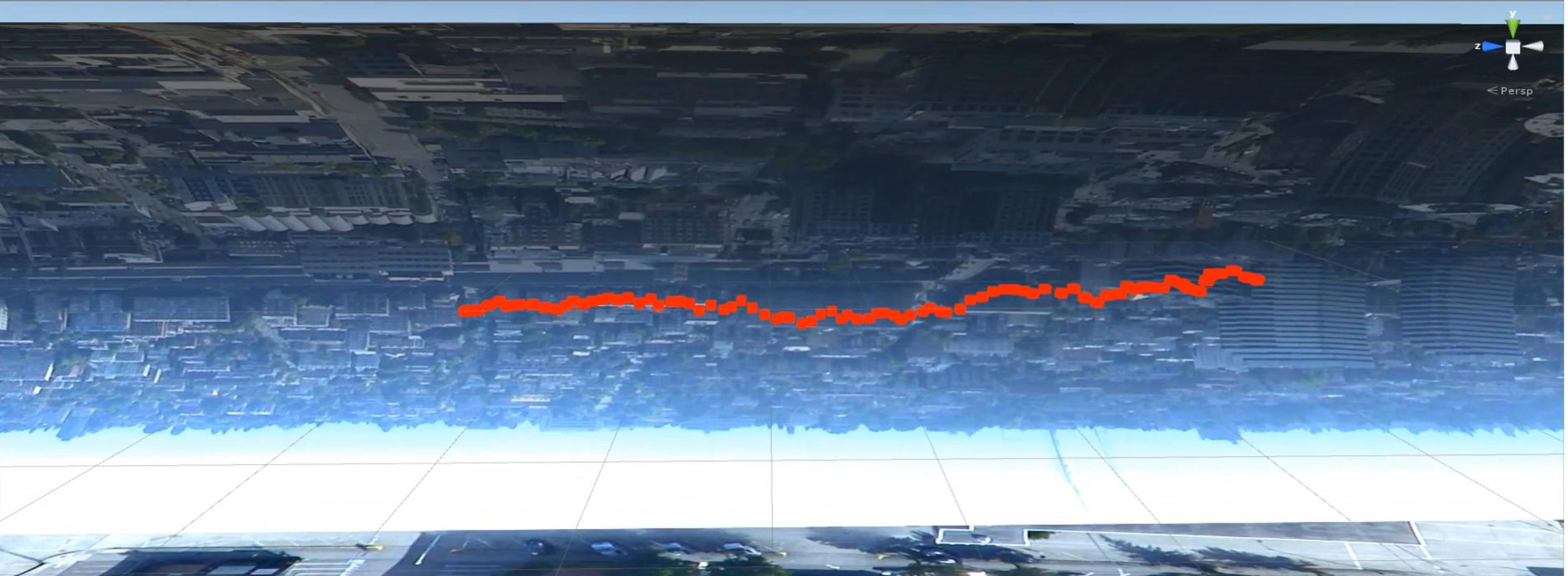}
  \caption{Simulated camera trajectory (shown in red).}
  	\end{subfigure}
	\begin{subfigure}[b]{0.225\textwidth}
		\includegraphics[width=\textwidth]{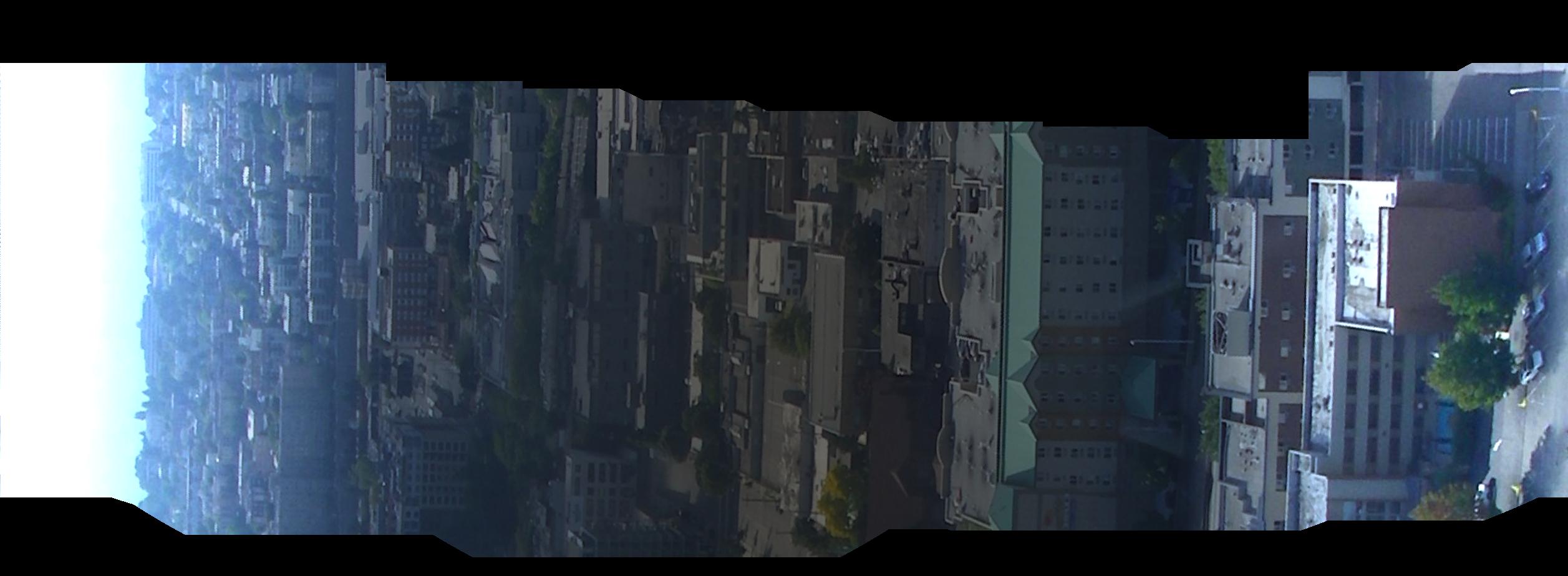}
  \caption{Ideal $3$D reconstruction using groundtruth camera pose.}
  	\end{subfigure}
	\begin{subfigure}[b]{0.225\textwidth}
		\includegraphics[width=\textwidth]{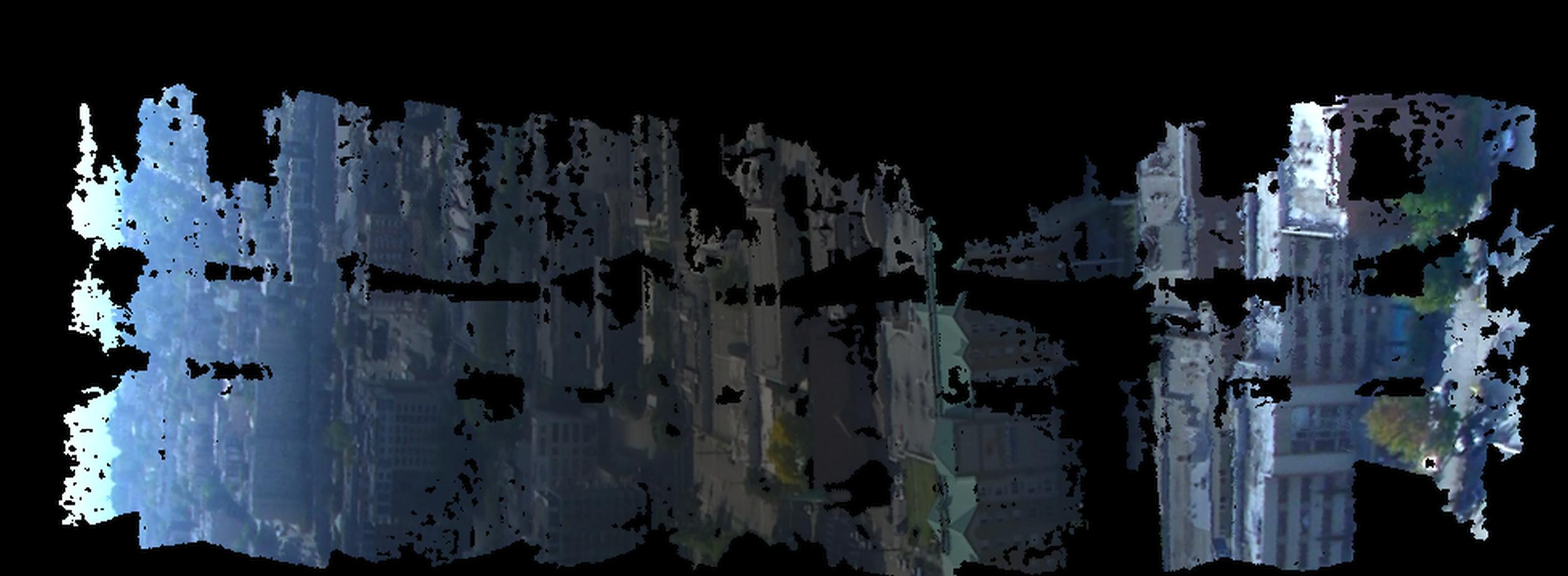}
  \caption{MVE $3$D reconstruction.}
  	\end{subfigure}
	\begin{subfigure}[b]{0.225\textwidth}
		\includegraphics[width=\textwidth]{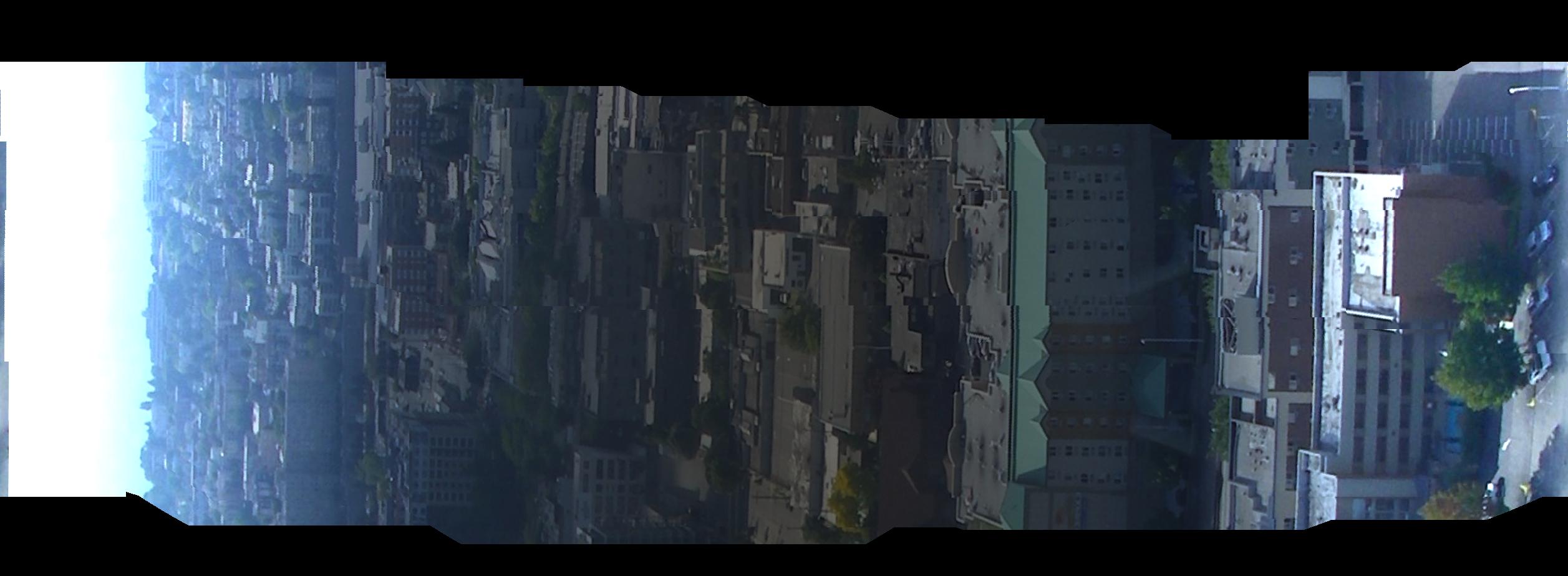}
  \caption{Our $3$D reconstruction.}
  	\end{subfigure}
  \caption{We add random jitters and movements to the simulated camera's trajectory during the image capturing process. MVE outputs a semi-dense reconstruction due to various areas lacking distinct feature points. Using Repmatch \cite{lin2016RepMatch} and dense reconstruction allow a more accurate and coherent cylindrical projection of the scene.
\label{fig::SfM_stitch_seattle} 
}
\end{figure}
\end{subsection}

\begin{subsection}{Real Data Results} \label{sec::real_datasets}
We use a lightweight GoPro HERO4 camera a TeraRanger-One range sensor to simultaneously capture an image and scalar depth information. We calibrate the GoPro to obtain its camera intrinsic and distortion parameters and take them into account during our pose estimation and reconstruction process.

The camera captures an image after rotating $36\degree$, which gives us ten images for every full $360\degree$ rotation. In lieu of reconstructing an underground tunnel using a UAV (confidential information), we demonstrate our imaging set-up in an underpass mounted on a tripod stand. The underpass has a flat horizontal floor, vertical walls on both sides and a cylindrical ceiling. We obtain a coarse geometry measurement to serve as a geometry prior for our algorithm. An example of the underpass is shown in Fig.~\ref{fig::underpass}(a).
\begin{figure}[t!]
\centering
\begin{subfigure}[b]{0.15\textwidth}
		\includegraphics[width=\textwidth]{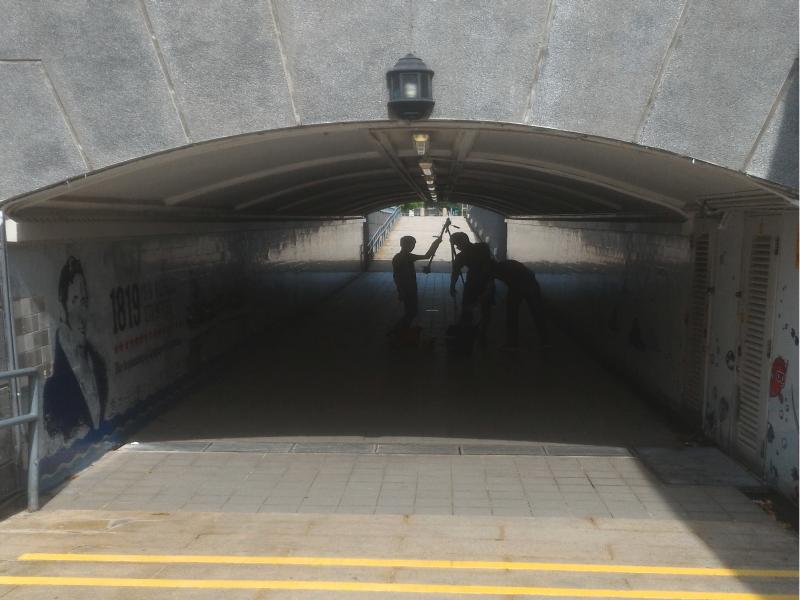} \caption{}
	\end{subfigure}  
	\begin{subfigure}[b]{0.15\textwidth}
		\includegraphics[width=\textwidth]{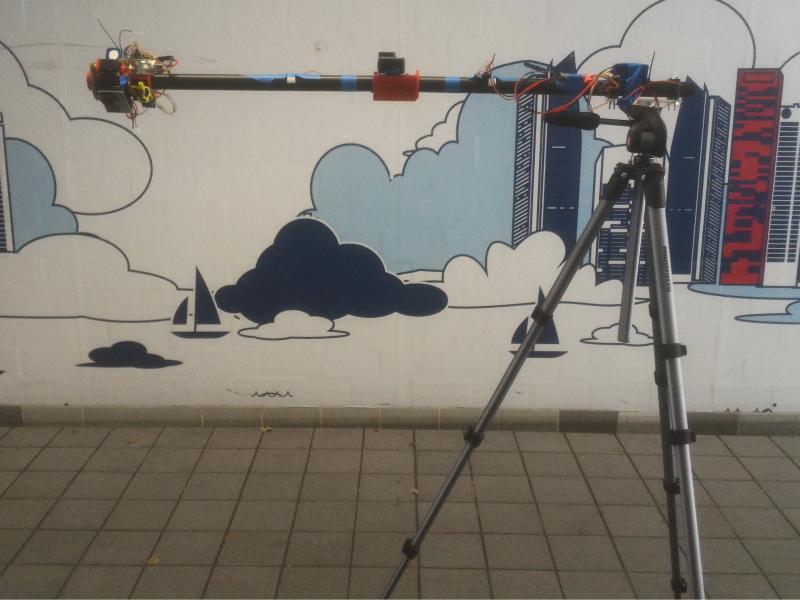} \caption{}
	\end{subfigure}  
	\begin{subfigure}[b]{0.15\textwidth}
		\includegraphics[width=\textwidth]{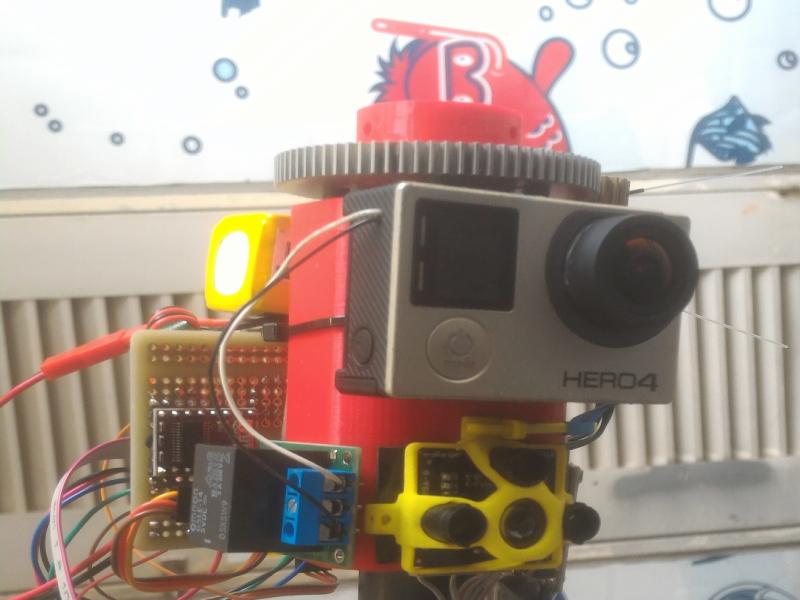} \caption{}
	\end{subfigure}  
  \caption{(a) Underpass used to capture our dataset. (b) Our camera setup. (c) Our camera rotating mechanism that closely simulates actual UAV's camera motion.}
\label{fig::underpass}
\end{figure}

We collect three different datasets using our set-up. In the first dataset, namely ``cylindrical dataset'', we rotate the camera to capture a panoramic $360\degree$ view of the scene before it is manually moved horizontally. In the second dataset, the tripod stand is moved manually after each image to obtain a ``spiral dataset''. Finally, in the third ``trolley dataset'', we placed the tripod stand on a trolley and move it manually while our rotating setup rotates the camera and captures images sequentially. Spiral and trolley datasets closely resemble actual data capturing process in the tunnel.  

\begin{subsubsection}{Cylindrical Dataset}
We capture seventeen complete rotations of the camera resulting in $170$ images. We use two different approaches to stitch the images after camera pose estimation. In the first approach, we identify which part of the underpass each pixel belongs to. Thereafter, it is projected onto a unit cylinder as discussed in \cite{raman_icis} which gives us a panoramic cylindrical view of the scene. The cyclic panoramic image is then wrapped around an internally hollow cylinder to display the panoramic view in Unity as shown in Fig.~\ref{fig::underpass_disp}. 

In the second approach, we utilize the known geometry of the scene to identify if a pixel in an image lies on the floor, right wall, left wall, or ceiling of the underpass and each section is stitched separately. The walls are shown separately for better view in Fig.~\ref{fig::underpass_cyl_spi}(a-b). The stitched images for the four sections are then textured onto four planes in Unity to obtain a $3$D representation of the scene as shown in Fig.~\ref{fig::mve_ours_results}. 
\end{subsubsection}
\begin{figure}[t!]
\centering
\begin{subfigure}[b]{0.225\textwidth}
		\includegraphics[width=\textwidth]{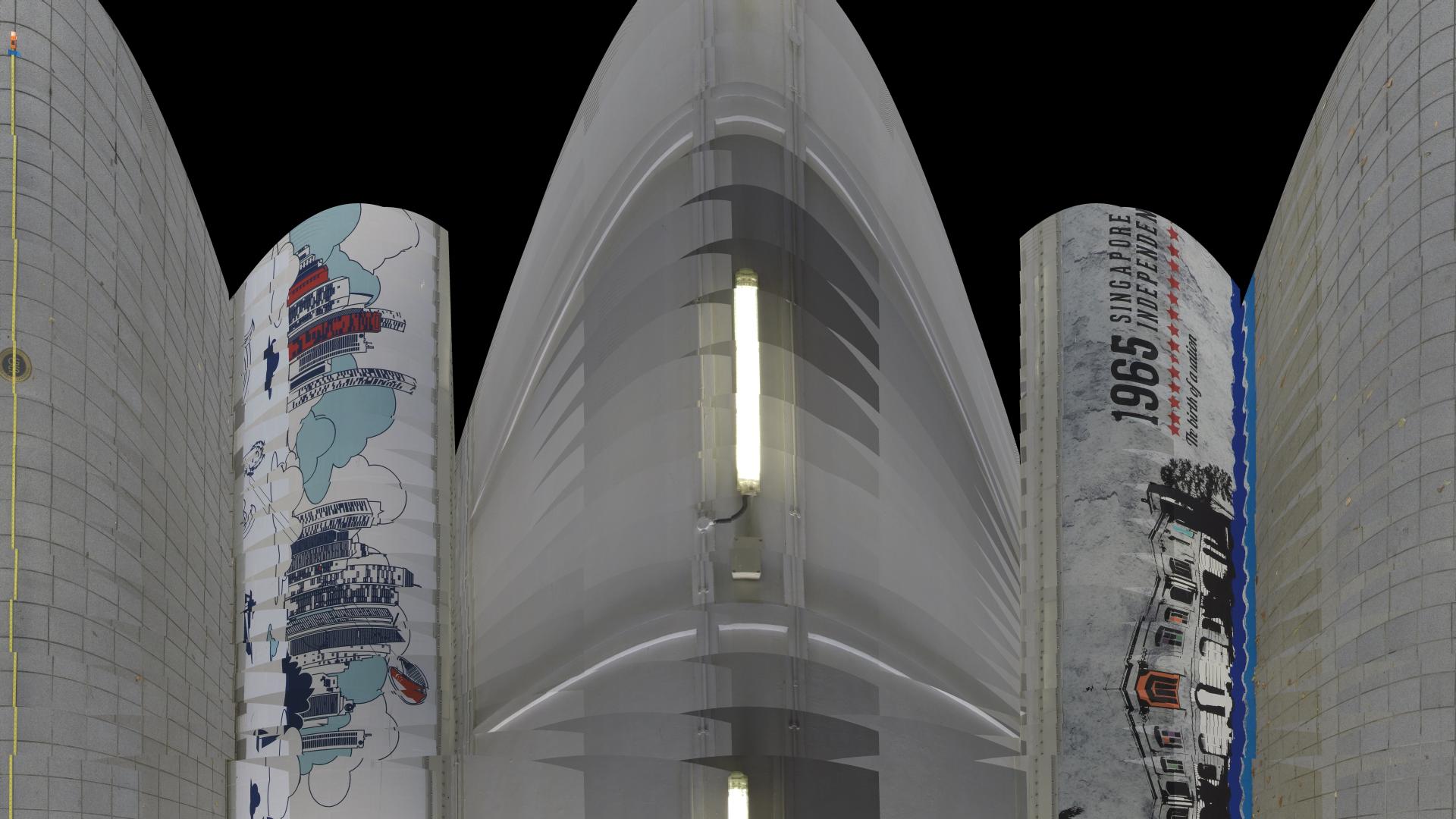}
		\caption{Cylindrical stitch.}
	\end{subfigure}  
	\begin{subfigure}[b]{0.225\textwidth}
		\includegraphics[width=\textwidth]{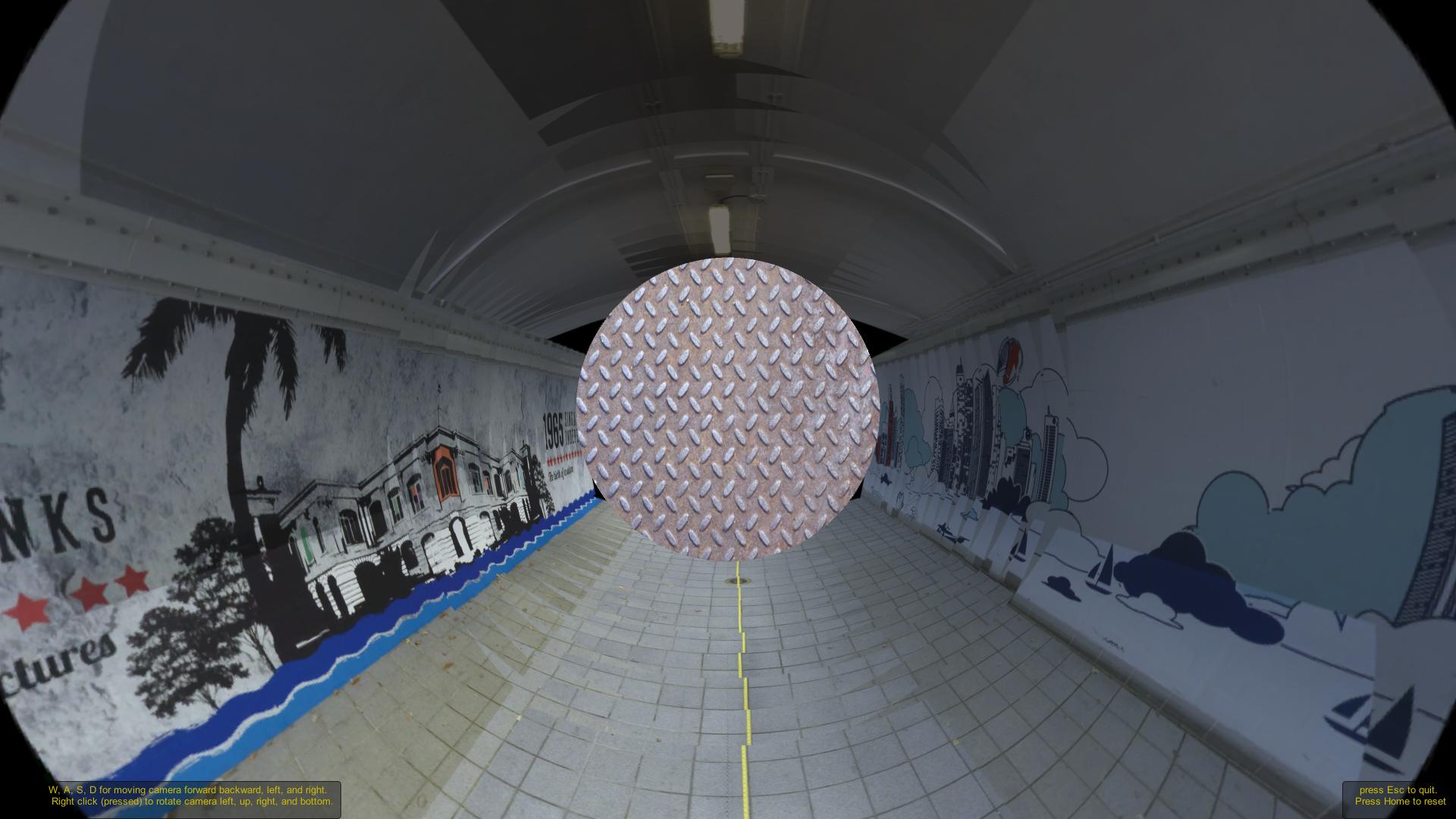}
		\caption{$3$D display in Unity.}
	\end{subfigure}   
  \caption{We display the underpass as a cylindrical projection in $2$D. The scene is rendered in a tunnel-like cylinder for visualization.}
\label{fig::underpass_disp}
\end{figure}
\begin{figure}[t!]
\centering
\begin{subfigure}[b]{0.225\textwidth}
		\includegraphics[width=\textwidth]{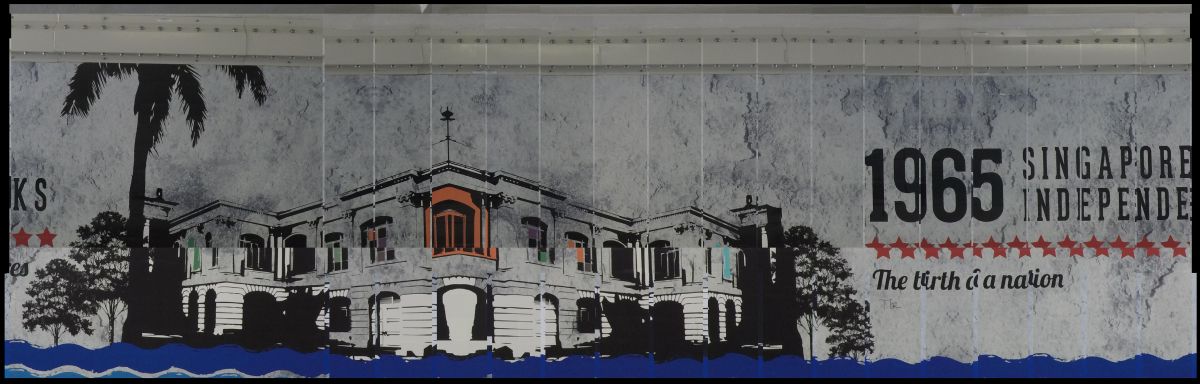}
		\caption{The Left wall of ``Cylindrical dataset''}
	\end{subfigure}  
	\begin{subfigure}[b]{0.225\textwidth}
		\includegraphics[width=\textwidth]{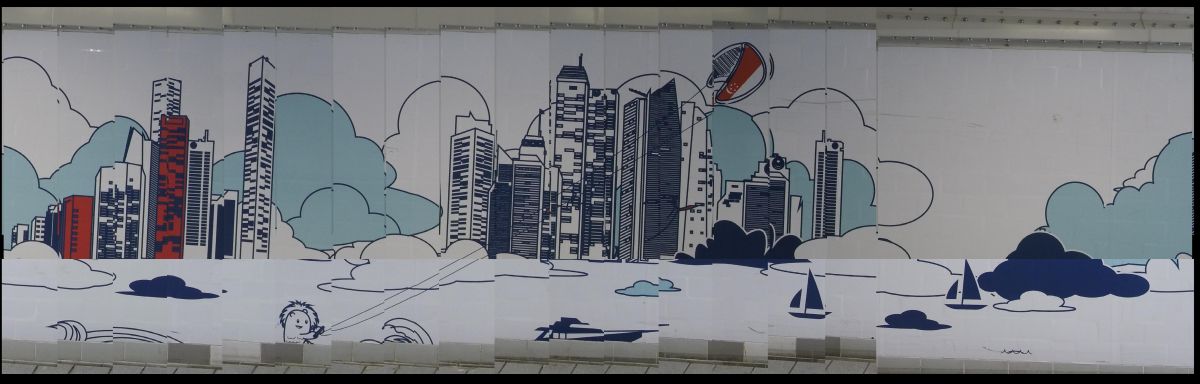}
		\caption{The Right wall of ``Cylindrical dataset''}
	\end{subfigure}  
  	\begin{subfigure}[b]{0.225\textwidth}
		\includegraphics[width=\textwidth]{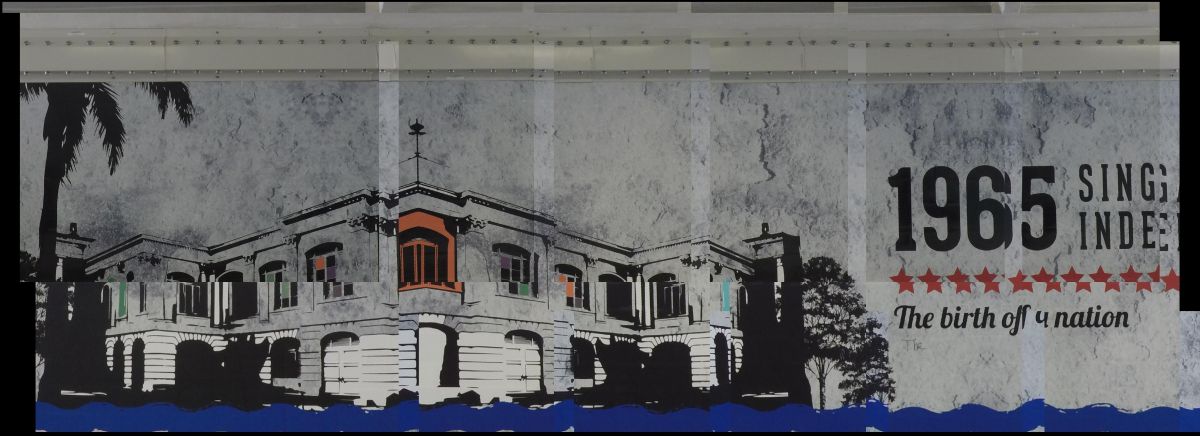}
		\caption{The Left wall of ``Spiral dataset''}
	\end{subfigure}  
	\begin{subfigure}[b]{0.225\textwidth}
		\includegraphics[width=\textwidth]{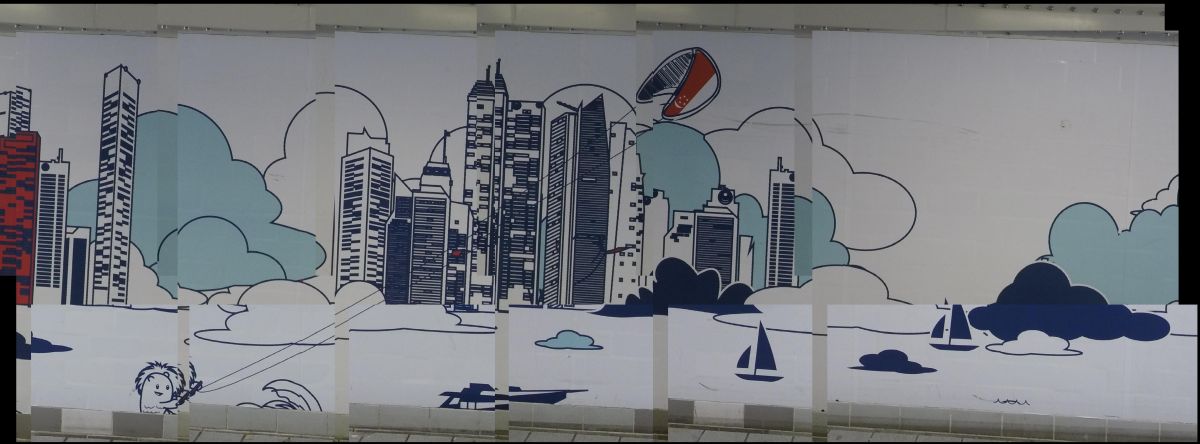}
		\caption{The Right wall of ``Spiral dataset''}
	\end{subfigure}  
	\begin{subfigure}[b]{0.45\textwidth}
		\includegraphics[width=\textwidth]{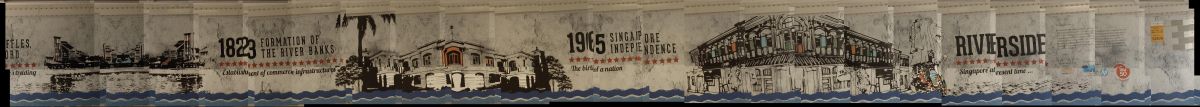}
		\caption{The Left wall of ``Trolley dataset''}
	\end{subfigure}  
	\begin{subfigure}[b]{0.45\textwidth}
		\includegraphics[width=\textwidth]{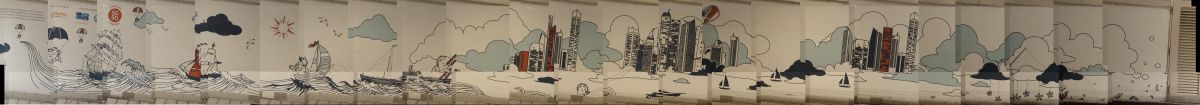}
		\caption{The Right wall of ``Trolley dataset''}
		\end{subfigure}  
  \caption{We display the $2$D textures obtained by our framework of the left and right walls for all three datasets. For the ``Cylindrical dataset'', the camera rotates rotates full $360 \degree$ before it is manually moved forward. For the ``Spiral dataset'' and ``Trolley dataset", manually moved forward per frame while the camera rotates $36 \degree$ automatically. }
\label{fig::underpass_cyl_spi}
\end{figure}
\vspace*{-0.4cm}
\begin{subsubsection}{Spiral Dataset}
In our second dataset, the camera is moved forward after every $36\degree$ of rotation, giving us a spiral image capture as shown in Fig.~\ref{fig::UAV_design_framework}(c). We capture fifty images for this dataset which corresponds to five full rotations and a forward movement of approximately $5$m in the underpass. 
Camera pose is estimated between consecutive frames as well as between every tenth frame ($1$ and $11$, $2$ and $12$ etc.) using RepMatch and further refined using our modified BA. 

Our experiments show that connections between every tenth frame and aggressive outlier removal is critical for good pose estimation. We stitch the images using the second approach discussed in cylindrical dataset section and Fig.~\ref{fig::underpass_cyl_spi}(c-d) shows the wall murals results.
\end{subsubsection}
\begin{subsubsection}{Trolley dataset}
In our third dataset, we use a trolley to traverse the entire underpass while the camera rotates and captures images. We capture $230$ images in this process while traversing approximately $22$m. 

We used the same processing pipeline as the second dataset. We observed that the lack of texture in the ceiling can result in noisy pose estimation. However our aggressive three-step pruning BA improves reconstruction results significantly. Figure~\ref{fig::floor_BA} shows the $3$D reconstruction of matching floor points with and without BA pruning to improve camera pose. Fig.~\ref{fig::underpass_cyl_spi}(e-f) shows result of the wall murals.
\end{subsubsection}
Figure~\ref{fig::mve_ours_results} shows a side-by-side visual comparison of $3$D reconstruction of the three datasets using our framework and MVE. We observe that MVE outputs a semi-dense pointcloud for Cylindrical dataset and a sparse pointcloud for both Spiral and Trolley datasets. Due to significant rotation and minimal overlap between images, MVE is unable to triangulate points in the latter two datasets. In comparison, our framework handles these challenges well and outputs a dense $3$D reconstruction as seen in Fig.~\ref{fig::mve_ours_results}(b). 

\begin{figure}[t!]
\centering
\begin{subfigure}[b]{0.225\textwidth}
		\includegraphics[width=\textwidth]{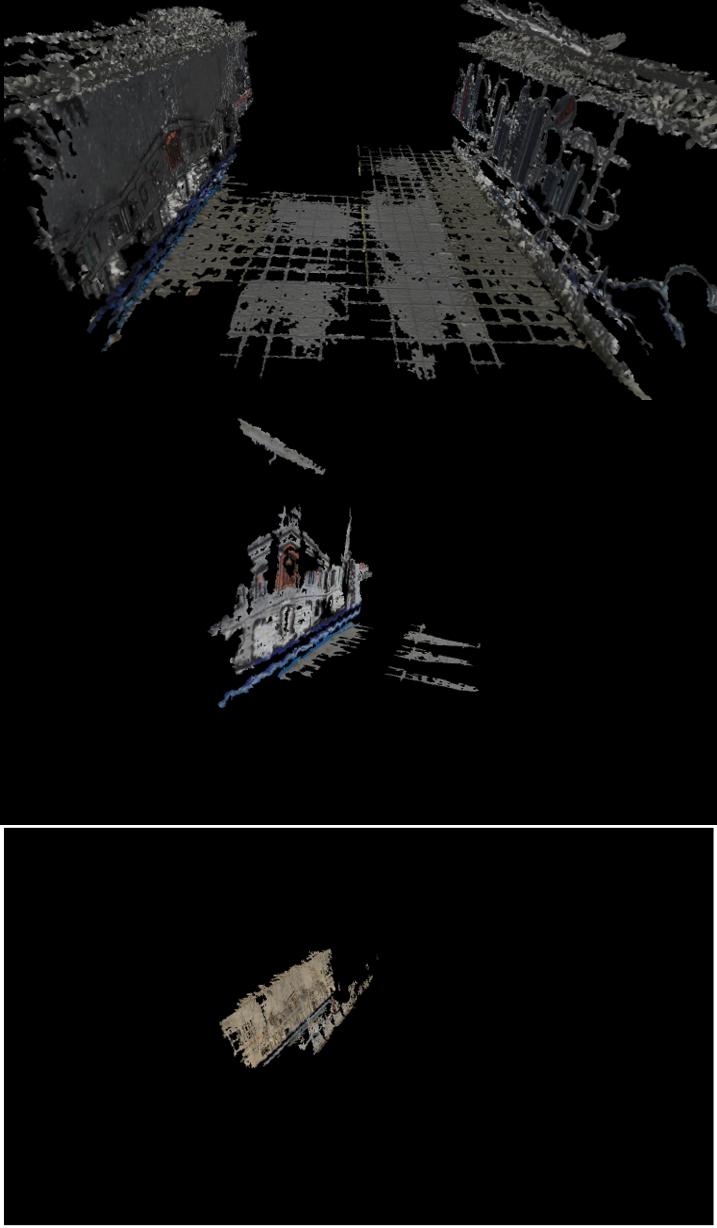}
\caption{3D Pointcloud obtained by MVE.}
\end{subfigure} 
\begin{subfigure}[b]{0.225\textwidth}
		\includegraphics[width=\textwidth]{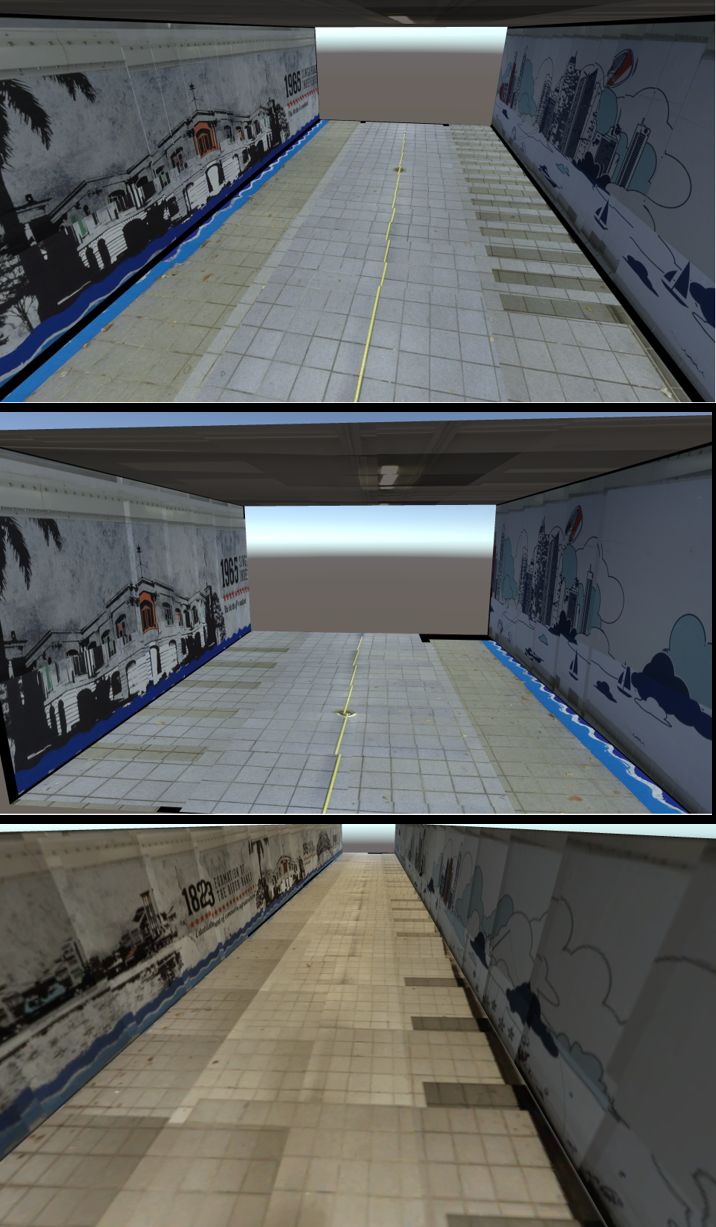}
\caption{Our computed pointcloud in Unity.}
\end{subfigure} 
  \caption{We compare our results with MVE \cite{fuhrmann2014mve} for Cylindrical, Spiral and Trolley dataset respectively. In Spiral and Trolley dataset, MVE fails to reconstruct majority of the scene due to significant rotational motion and lack of distinct feature points on the ceiling. Our framework returns a dense $3$D reconstruction of the scene using a handful of images. We texture-map planes in Unity to show our dense reconstruction of walls, floor and ceiling for visualization.} 
\label{fig::mve_ours_results}
\end{figure}
\end{subsection}

\end{section}

\begin{section}{Conclusion}\label{sec::conclusion}
We presented a system that fully reconstructs the given scene using a spirally moving camera and displays the $3$D scene in Unity for an interactive display. The presented method excels in scenes where prior geometrical information is available. Our framework is capable of handling significant rotational motion and  reconstructing the scene with few images in contrast with current state-of-the-art techniques. 

In the future, we intend to infer scene geometry using neural networks in place of providing a geometry prior \cite{itsc_2019_raman}. In addition, a seam-carving approach can be used to compensate for exposure differences between consecutive images. Alterntaively, a better designed UAV \cite{8611114, 102166, 8452686} can be used to make the data capturign task less challenging.
\end{section}

\begin{section}{Acknowledgement}\label{sec::ack}
This work was partially supported by the National Research Foundation, Prime Minister’s Office, Singapore under its Environment and Water Research Programme (Project Ref No. 1502-IRIS-03). This programme is administered by PUB, Singapore's national water agency.
\end{section}

\bibliographystyle{IEEEtran}
\bibliography{iros2019_raman}

\end{document}